\documentclass[article,journal]{IEEEtran}

\usepackage{amsmath,amssymb,amsfonts}
\usepackage[linesnumbered,ruled]{algorithm2e}
\usepackage{listings}
\usepackage{float}
\usepackage{colortbl}
\usepackage{orcidlink}

\usepackage{multirow}
\usepackage{amsthm}
\usepackage{mathrsfs}
\usepackage{xcolor}
\usepackage{manyfoot}
\usepackage{booktabs}
\usepackage{bbm}
\usepackage{mathtools}
\usepackage{tikz}
\usepackage{threeparttable}
\usepackage{caption}

\usepackage{array}
\usepackage[caption=false,font=footnotesize,labelfont=rm,textfont=rm]{subfig}
\usepackage{textcomp}
\usepackage{stfloats}
\usepackage{url}
\usepackage{verbatim}
\usepackage{graphicx}

\usepackage{balance}
\usepackage{nomencl}
\usepackage[bottom]{footmisc}
\usepackage{hyperref}

\def\BibTeX{{\rm B\kern-.05em{\sc i\kern-.025em b}\kern-.08em
    T\kern-.1667em\lower.7ex\hbox{E}\kern-.125emX}}

\theoremstyle{remark}

\makenomenclature

\usepackage{stackengine}

\newcommand{\replacename}[1]{\textcolor{red}{[Hidden names]}}
\newcommand{\replaceaffiliation}[1]{\textcolor{red}{[Hidden affiliations]}}

\newcommand{\footurl}[2]{\footnote{\href{#1}{#2: \texttt{#1}}}}

\begin{document}
\title{Distributed Relative Localization for Homogeneous Multi-Robot Systems through UWB Ranging and Limited Communications}

\author{
~~~Zhiqiang Cao\textsuperscript{\orcidlink{0000-0002-0823-7155}},
Ran Liu\textsuperscript{\orcidlink{0000-0002-6343-4645}}, \IEEEmembership{Senior Member, IEEE}, 
Billy Pik Lik Lau\textsuperscript{\orcidlink{0000-0001-5133-2791}}, \IEEEmembership{Senior Member, IEEE}, \\
Chau Yuen\textsuperscript{\orcidlink{0000-0002-9307-2120}}, \IEEEmembership{Fellow, IEEE},
U-Xuan Tan\textsuperscript{\orcidlink{0000-0002-5757-1379}}, \IEEEmembership{Senior Member, IEEE}

\thanks{

Z. Cao and U-X. Tan are with the Engineering Product Development, Singapore University of Technology and Design, Singapore 487372. zhiqiang\_cao@mymail.sutd.edu.sg and uxuan\_tan@sutd.edu.sg.

R. Liu, B. P. L. Lau, and C. Yuen are with the School of Electrical and Electronic Engineering, Nanyang Technological University, Singapore 639798. \{ran.liu, billy.laupl, chau.yuen\}@ntu.edu.sg. 

(Corresponding author: Chau Yuen)
}
}

\markboth{IEEE PREPRINT, ~2025}%
{How to Use the IEEEtran \LaTeX \ Templates}

\maketitle

%%%%%%%%%%%%%%%%%%%%%%%
%%%%%% Abstract %%%%%%%
%%%%%%%%%%%%%%%%%%%%%%%
\begin{abstract}
Accurate and reliable relative localization is crucial for multi-robot applications like exploration, search, and rescue missions. 
LiDAR-based solutions offer high accuracy in localizing surrounding objects; however, distinguishing homogeneous robots with similar appearances remains challenging due to the lack of distinctive identification features. 
In this paper, we propose a fully distributed relative pose estimation approach by integrating LiDAR, UWB, and odometry measurements, allowing each robot to accurately and continuously localize its teammates without external infrastructure. Specifically, potential anonymous teammate robot clusters from LiDAR scans are tracked by a dynamic tracker. We then identify teammate robots from these tracked anonymous clusters using a joint matching strategy, ensuring reliable data association between robots and clusters. 
Finally, by combining the corresponding LiDAR observations, UWB ranging, and odometry measurements, each robot precisely localizes others while minimizing data exchange. The system requires only odometry data exchange through onboard UWB, eliminating the need for additional communication infrastructure like WiFi routers or mesh networks. 
Extensive simulation and real-world experiments demonstrate the effectiveness and reliability of the proposed relative localization approach.

%Experiments with three robots demonstrate that our approach achieves an average relative translational accuracy of 0.123m and rotational accuracy of 2.986$^{\circ}$, with a data exchange of 0.48KB/s among robots, representing a substantial improvement over the state-of-the-art approach.

\end{abstract}


%% =================== %%
%%  Introduction       %%
%% =================== %%

\section{Introduction}\label{section-introduction}
\IEEEPARstart{M}{ulti-robot} systems have gained significant interest in the robotics research community over the past decades due to their superior applicability and efficiency compared against single robot systems in various applications, such as exploration, search, and rescue missions \cite{Introduction-multi-robot-system-1}-\cite{Introduction-multi-robot-exploration}.
To execute these multi-robot cooperative tasks in unknown scenarios, each robot must be capable of reliably and accurately estimating the relative position of others with respect to itself \cite{Introduction-relative-localization-1}-\cite{Introduction-relative-localization-3}.
This is essential for maintaining formation, preventing collisions, and coordinating actions seamlessly.

To date, the majority of multi-robot research depends on external sensing devices, such as global positioning systems (GPS) \cite{Introduction-GPS}, motion capture systems \cite{Introduction-motion-caption}, and infrastructure-based Ultra-WideBand (UWB) \cite{Introduction-UWB-positioning}, to achieve relative localization between robots.
Although these mature techniques can provide precise localization, they require information fusion with a centralized system, making them sensitive to single-point-of-failure.
In addition, GPS signals are often impeded by structural obstructions and susceptible to multipath effects, limiting its effectiveness in indoor environments.
Both motion capture systems and infrastructure-based UWB positioning systems require the installation of external anchors, which restrict their applicability in unknown environments.

\begin{figure}[t]
\centering
\includegraphics[width=90mm]{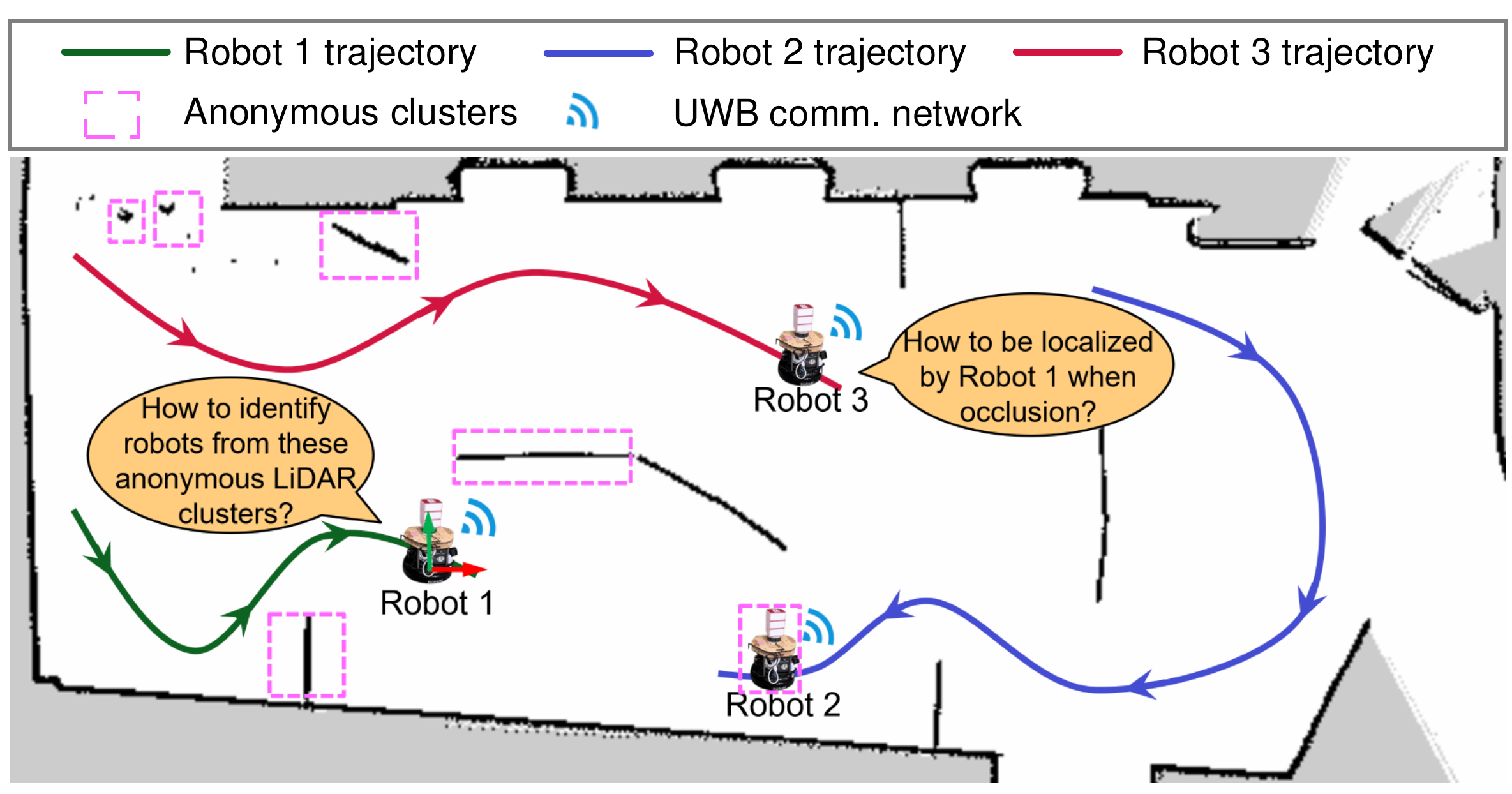}
\caption{\small Illustration of the relative localization problem considered in this paper, from Robot 1's perspective.
Robot 1 attempts to localize others in its local frame under the following conditions: unknown initial pose correspondence, anonymous LiDAR clusters, and possible occlusions during movement.
Additionally, data transmission relies solely on the UWB communication network, mounted on the robot, without any external communication devices.
Each robot in the multi-robot system independently executes the same relative localization procedure.}
\label{figure-section1-overview}
\end{figure}

\vspace{-4pt}
Recently, LiDAR and visual odometry techniques have been widely studied \cite{Introduction-LiDAR-visual-odometry-1}-\cite{Introduction-LiDAR-visual-odometry-3} due to their capability to enable infrastructure-free self-localization in unknown environments.
However, unknown initial pose correspondence among robots raises a challenge for precise relative localization in multi-robot systems.
Moreover, pose drift is inevitable during long-time running, and it tends to be more severe in 2D LiDAR-based configurations due to the sparsity of planar geometric features.
Although SLAM techniques can correct drift through loop closure, this mechanism is not always available in weakly textured environments or before the robot revisits previously explored areas.
Another well-known relative localization approach is matching common features in LiDAR scans or camera images, shared between robots.
This approach demands high computational resources and communication bandwidth, making it impractical in communication-constrained and weak-textured environments.
Direct robot-to-robot observations using LiDAR and cameras offer another solution for relative position estimation.
However, these sensors encounter difficulties with anonymous detection results due to the similar structure of objects (i.e., data association problem) and their dependence on line-of-sight (LOS) conditions \cite{Introduction-anonymous-detection-1}\cite{Introduction-anonymous-detection-2}.
In contrast, wireless sensors, like UWB, demonstrate greater resilience to non-line-of-sight (NLOS) situations.
Meanwhile, it is able to provide accurate range measurements with unique identification, effectively eliminating data association ambiguity \cite{Introduction-UWB-properties}\cite{Introduction-UWB-properties-2}.
Furthermore, UWB supports dual functionality, serving as a communication module for transmitting a small amount of data to the surrounding UWB nodes while concurrently performing ranging \cite{Introduction-UWB-data-transmission}.
Nonetheless, the practicality of UWB in relative localization is limited due to the lack of bearing information among UWB nodes.

Leveraging the complementary strengths of odometry, LiDAR, and UWB, we propose a fully distributed relative localization approach for a homogeneous multi-robot system that operates effectively under challenging conditions, as illustrated in Fig. \ref{figure-section1-overview}.
Particularly, UWB ranging and LiDAR clusters generated by the adaptive clustering algorithm \cite{Methodology-adaptive-clustering} are first processed to eliminate outliers.
Then, a dynamic tracker is designed to track these filtered anonymous LiDAR clusters within the robot's own global frame.
Subsequently, we identify teammate robots from these tracked LiDAR clusters based on a joint matching strategy.
Once the teammate robot is identified and meets the cumulative displacement criteria, a separate teammate tracker is assigned for continuous relative pose estimation.
All communication demands in the proposed system are fulfilled by the tailored UWB communication network, without external communication devices.
In summary, the contributions of this paper are as follows:
\begin{itemize}
    \item We propose a robust teammate identification approach based on a joint matching strategy, ensuring reliable data association between teammate robots and LiDAR clusters.
    Our approach eliminates the need for appearance-based modification, such as the addition of visual markers or other physical alterations, to identify the robots, thereby enhancing the scalability of multi-robot systems.
    
    \item We present a fully distributed relative localization system that fuses LiDAR, odometry, and UWB ranging measurements without requiring external infrastructure.
    The system is built upon the UWB communication network used to transmit odometry data among robots, with each message containing only 8 Bytes per robot, while ensuring reliable and efficient localization.
    
    \item {To evaluate the performance of the proposed approach, we conducted several experiments in simulation and real-world environments. The results demonstrate that our approach achieved high relative localization accuracy while incurring minimal data exchange, limited to 0.48 KB/s at a transmission frequency of 20 Hz in a three-robot system.}
\end{itemize}

The rest of this paper is structured as follows:
The related works are introduced in Section \ref{section-related-works}.
Subsequently, we present the proposed distributed relative localization system in Section \ref{section-methodology}.
Section \ref{section-experiments} shows the experimental setups and results.
Finally, this paper is concluded and future works are discussed in Section \ref{section-conclusion-and-future-works}.

\begin{table}[t]
\centering
\setlength{\abovecaptionskip}{0pt}%    
\setlength{\belowcaptionskip}{0pt}%
\caption{\small Comparison of related works on relative localization in multi-robot systems.}
\label{table-related-work}
\resizebox{.49\textwidth}{!}{
\begin{tabular}{|c|c|c|c|c|}
\hline
\textbf{Work} & \textbf{Sensors} & \textbf{Markers} & \textbf{Data Exchanged} & \textbf{Initial Pose} \\ 
\hline\hline  % Double line for better separation
\cite{Related-work-Lajoie} & \begin{tabular}[c]{@{}c@{}}LiDAR +\\ Camera + IMU\end{tabular} & N/A & \begin{tabular}[c]{@{}c@{}}Ego state,\\ Feature descriptors\end{tabular} & Unknown \\ \hline
\cite{Related-work-Weinstein}\cite{Related-work-Lusk} & Camera + IMU & N/A & Ego state & Known \\ \hline
\cite{Introduction-anonymous-detection-1} & LiDAR + IMU & Reflective taps & \begin{tabular}[c]{@{}c@{}}Ego state,\\ Mutual observations\end{tabular} & Unknown \\ \hline
\cite{Related-work-xun} & \begin{tabular}[c]{@{}c@{}}Camera + UWB +\\ IMU + IR LEDs\end{tabular} & LED coded boards & \begin{tabular}[c]{@{}c@{}}IMU,\\ Mutual observations\end{tabular} & Unknown \\ \hline
\cite{Related-work-cao-subgraph-matching} & LiDAR + UWB & N/A & UWB ranging meas. & Unknown \\ \hline
\cite{Related-work-cao}\cite{Related-work-Fishberg} & Four UWBs & N/A & N/A & Unknown \\ \hline
\cite{Related-work-Liu} & UWB + Odom. & N/A & Odometry & Unknown \\ \hline
\cite{Introduction-anonymous-detection-2} & \begin{tabular}[c]{@{}c@{}}Camera + \\ IMU + UWB\end{tabular} & N/A & \begin{tabular}[c]{@{}c@{}}Ego state, Visual obs.,\\ Map-based meas.\end{tabular} & Unknown \\ \hline
\textbf{Ours} & \begin{tabular}[c]{@{}c@{}}LiDAR + UWB +\\ Odom.\end{tabular} & N/A & Odometry & Unknown \\ \hline
\end{tabular}}
% \begin{tablenotes}
% \scriptsize
% \item[*] \textbf{Markers} stands for tools used to facilitate LiDAR or camera detection.
% \item[$\dagger$] \textbf{Initial pose} denotes the initial relative pose among robots.
% \end{tablenotes}
\end{table}

\section{Related Works}\label{section-related-works}
Relative localization for a multi-robot system operating in GPS-denied and communication-constrained environments without any external infrastructure has emerged as a significant research focus in the robotics community.
This section covers the existing studies on relative pose estimation in multi-robot systems based on LiDAR, cameras, and wireless sensors.

In recent years, LiDAR-based and visual-based solutions have gained popularity in the robotics community.
Lajoie \textit{et al.} \cite{Related-work-Lajoie} presented a two-stage matching approach for constructing inter-robot loop closure (i.e., relative pose between robots) by exchanging the features extracted from the image and LiDAR point cloud among robots.
However, a substantial communication load is needed for exchanging features, and feature matching requires high computational resources.
To relieve communication and computational load, Weinstein \textit{et al.} \cite{Related-work-Weinstein} and Lusk \textit{et al.} \cite{Related-work-Lusk} acquire the relative pose between robots by sharing the robot's ego-state generated by Visual-Inertial Odometry (VIO) \cite{Related-work-VIO}.
These methods assume a known initial pose correspondence among robots and are prone to drift over time, which is an inherent limitation in VIO.

In addition, LiDAR-based or visual-based object detection methods have emerged as promising alternatives for accurate relative position estimation.
However, associating the detection results with specific teammate robots remains a challenge for a group of homogeneous robots due to their similar structures or appearances.
Zhu \textit{et al.} \cite{Introduction-anonymous-detection-1} addressed data association by implementing trajectory matching between the tracked anonymous LiDAR clusters and odometry trajectories received from other robots, in which each robot was attached with reflective taps for LiDAR detection.
Xun \textit{et al.} \cite{Related-work-xun} used active infrared LEDs to encode drone ID, allowing for straightforward identification of each teammate drone.
These methods require customized platforms, which may restrict their applications in practical deployment.
Cao \textit{et al.} \cite{Related-work-cao-subgraph-matching} addressed the data association problem between teammate robots and LiDAR clusters through a subgraph matching approach.
Although this approach requires little data exchange and performs effectively in normal scenarios, it can fail when robots form completely symmetrical structures and is computationally inefficient.
Moreover, LiDAR or visual-based solutions may be limited by LOS conditions.
In this paper, the data association problem can be solved by the proposed joint matching strategy without necessitating any changes to the robot's structure and visual appearance, while the continuity of the state estimation is kept by efficient fusion of multiple sensing modalities data.

Compared to LiDAR and visual camera, wireless sensors such as UWB can offer ID information, thereby avoiding the data association problem, and are suitable in NLOS scenarios.
Cao \textit{et al.} \cite{Related-work-cao} proposed a relative localization method only relying on several UWB nodes equipped on the robot.
Subsequently, Fishberg\textit{et al.} \cite{Related-work-Fishberg} improved its performance through an advanced optimization method.
Although fully distributed, these methods have limited accuracy and are unsuitable for small robots due to their UWB node array requirement.
Liu \textit{et al.} \cite{Related-work-Liu} fused short-term UWB ranging and odometry measurements to achieve relative localization between robots.
However, this method requires multiple samples to generate a coarse initial guess for subsequent non-linear optimization.
Xu \textit{et al.} \cite{Introduction-anonymous-detection-2} integrated UWB ranging, IMU, map-based measurements, and visual detections using YOLOv3-tiny \cite{Related-work-yolo} to enhance the performance of localization.
Nonetheless, this approach requires broadcasting a large volume of information, including ego state, landmark feature descriptors, and visual detections, which exceeds the maximum communication capacity of UWB.
Moreover, visual detection is a learning-based method, which typically requires additional costs to prepare the training dataset.
In this paper, we propose an out-of-the-box method that requires neither data labeling nor customizing platforms, which effectively resolves the data association problem and provides a precise initial guess for subsequent nonlinear optimization.
Further details are provided in Section \ref{section-methodology}.
Table \ref{table-related-work} presents a comprehensive comparison of relevant studies in the field of relative localization for multi-robot systems.

\begin{figure*}[t]
\vspace{-\baselineskip}
\centering
\setlength{\abovecaptionskip}{0pt}%    
\setlength{\belowcaptionskip}{-10pt}%
\includegraphics[width=130mm]{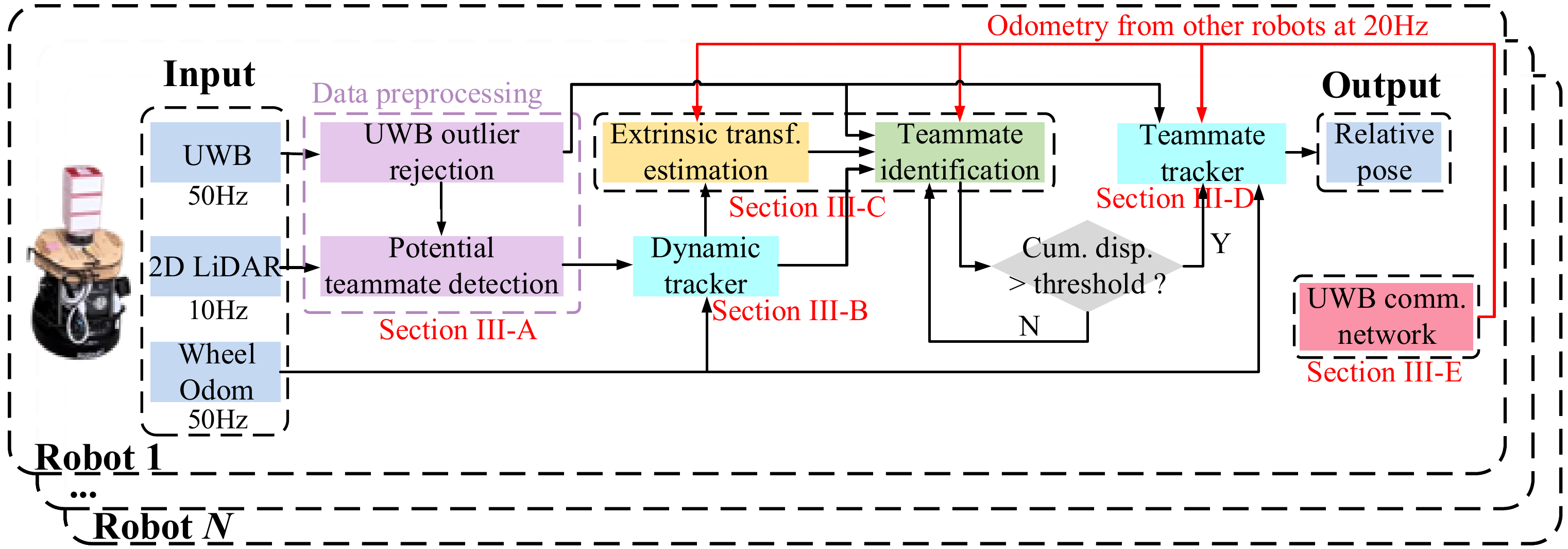}
\caption{\small Architecture of the proposed fully distributed relative localization approach, comprising five core modules.
The data preprocessing module eliminates outlier UWB and LiDAR observations.
The dynamic tracker module tracks the anonymous LiDAR clusters.
The extrinsic transformation estimation and teammate identification module addresses the unknown initial pose correspondence and data association problem.
The teammate tracker module enables continuous localization of teammate robots by fusing UWB, LiDAR, and odometry data.
All the communications among robots rely solely on the UWB communication network module.
}
\label{figure-sectionIII-flowchart}
\end{figure*}

\section{Proposed relative localization framework}\label{section-methodology}
In this section, we introduce key notations and problem formulation studied in this paper.
Following this, we describe the proposed relative localization framework, involving five key components:
(i) \textit{Data preprocessing} in Section \ref{subsection-data-preprocessing}, which rejects the abnormal UWB ranging measurements and LiDAR clusters,
(ii) \textit{Dynamic tracker} in Section \ref{subsection-temporary-tracker}, which tracks the filtered anonymous LiDAR clusters within the robot's own global frame,
(iii) \textit{Extrinsic transformation estimation} and \textit{teammate identification} in Section \ref{subsection-global-extrinsic-estimation-and-teammate-identification}, which address the unknown initial pose correspondence problem by estimating the transformation between trajectories and identifying teammate robots,
(iv) \textit{Teammate tracker} in Section \ref{subsection-teammate-tracker}, capable of continuously localizing teammate robots within the robot's own global frame,
and (v) \textit{UWB communication network} in Section \ref{subsection-uwb-communication-network}, designed to broadcast the odometry measurements to nearby robots within communication range.
An overview of the proposed system framework is illustrated in Fig. \ref{figure-sectionIII-flowchart}.

Let us consider a multi-robot system of $N$ robots with similar appearance, each one carries a LiDAR, UWB, and wheel odometry.
We assign an index $i$ to each robot, which is consistent with the UWB ID on the robot, $i\in \left\{ 1, 2, \cdots, N\right\}$.
Each robot has its own global frame $G_{i}$ and local frame $b_{i}$.
At a given time $t$, Robot $i$ can observe the ego motion state, which is represented by a $SE(2)$ transformation matrix, i.e., ${^{G_{i}}\mathbf{x}_{i}^{t}}=({^{G_{i}}\mathbf{p}_{i}^{t}}, {^{G_{i}}\mathbf{R}_{i}^{t}})$, based on odometry measurements in its own global frame $G_{i}$. ${^{G_{i}}\mathbf{p}_{i}^{t}}=\left [ {^{G_{i}}x_{i}^{t}}, {^{G_{i}}y_{i}^{t}} \right ]\in \mathbb{R}^{2}$ is the translation vector and ${^{G_{i}}\mathbf{R}_{i}^{t}}\in SO(2)$ refers to the rotation matrix of heading $^{G_{i}}\theta _{i}^{t}$.
The raw UWB ranging measurement between Robot $i$ and Robot $j$ at time $t$, measured by Robot $i$, is represented as $r_{i,j}^{t}$, $i,j\in \left\{ 1, 2, \cdots , N\right\}$ and $i\neq j$, while $\check{r}_{i,j}^{t}$ denotes the filtered UWB ranging measurement.
${^{b_{i}}\mathbf{C}^{t}}=\left\{{^{b_{i}}\mathbf{c}_{1}^{t}}, {^{b_{i}}\mathbf{c}_{2}^{t}}, \cdots, {^{b_{i}}\mathbf{c}_{m}^{t}} \right\}$ indicates the set of LiDAR clusters generated by Robot $i$ at time $t$, where ${^{b_{i}}\mathbf{c}_{m}^{t}}\in \mathbb{R}^{2}$ represents the position of the $m$-th LiDAR cluster in Robot $i$'s local frame $b_{i}$.
The set of filtered LiDAR clusters is denoted as $^{b_{i}}\check{\mathbf{C}}^{t}$.
The relative pose of Robot $j$ with respect to Robot $i$ at time $t$ in Robot $i$'s local frame is represented as $^{b_{i}}\mathbf{x}_{i,j}^{t}$, thus the pose of Robot $j$ in Robot $i$'s global frame can be represented as $^{G_{i}}\mathbf{x}_{j}^{t}={^{G_{i}}\mathbf{x}_{i}^{t}}\times {^{b_{i}}\mathbf{x}_{i,j}^{t}}$.
For clarity, the remainder of this section is presented from the perspective of Robot $i$.

Note that throughout this paper, no universal world global frame is assumed for the multi-robot system.
Each robot defines its own global frame, fixed at its initial point, while the local frame dynamically follows the robot's motion.
Moreover, the initial pose correspondence among robots is unknown.

\vspace{-10pt}
\subsection{Data Preprocessing}\label{subsection-data-preprocessing}
To ensure that our proposed method is not affected by severely noisy measurements, we perform data pre-processing prior to the implementation of the main algorithm.

\subsubsection{UWB outlier rejection}\label{subsubsection-UWB-outlier-rejection}
In practical applications, raw UWB ranging measurements are often susceptible to multipath effects and electromagnetic interference, especially in complex environments.
This results in unpredictable outliers, degrading the system's overall performance.
Thus, effectively recognizing and removing these outliers is essential.

\begin{figure}[t]
\vspace{-\baselineskip}
\centering
\setlength{\abovecaptionskip}{0pt}%    
\setlength{\belowcaptionskip}{0pt}%
\subfloat[Ranging from Robot 1 to Robot 2]{
    \includegraphics[width=42mm]{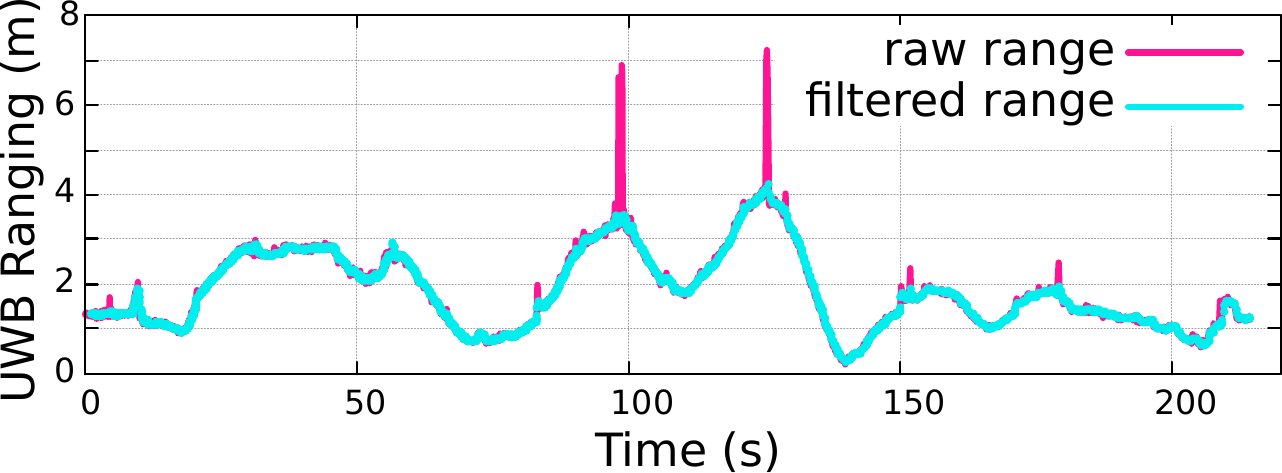}
    \label{uwb-outlier-rejection-1}
}
\subfloat[Ranging from Robot 1 to Robot 3]{
    \includegraphics[width=42mm]{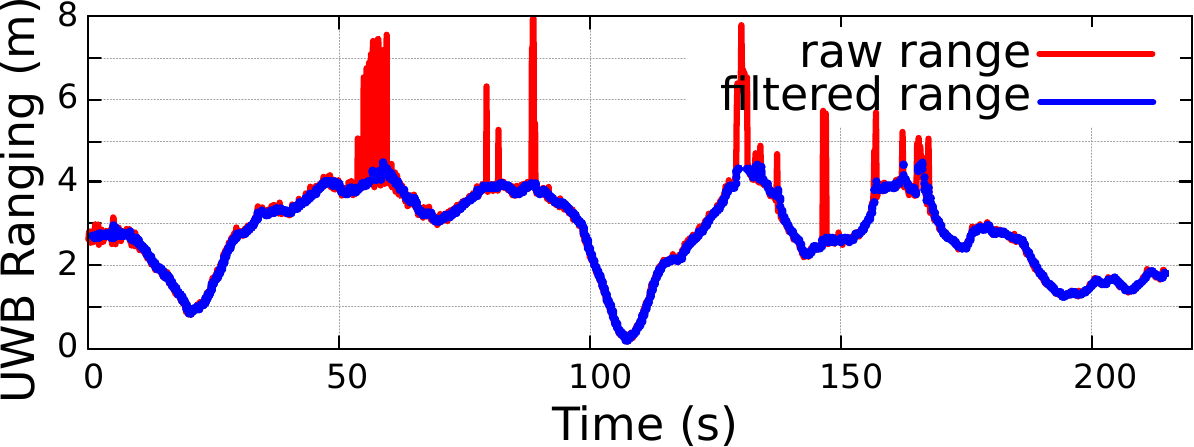}
    \label{uwb-outlier-rejection-2}
}
\caption{\small Illustration of raw and filtered UWB ranging data, as measured by Robot $1$. (a) UWB ranging measurements of Robot $2$. (b) UWB ranging measurements of Robot $3$.}
\label{figure-uwb-outlier-rejecetion}
\end{figure}

Each robot moves continuously in an environment, the UWB ranging measurements at two consecutive timestamps, $r_{i,j}^{t-1}$ and $r_{i,j}^{t}$, should not exhibit large variation.
The ranging measurements are expected to remain smooth over time.
To enhance measurement reliability, we employ a low-pass filter \cite{Digital-signal-processing} to smooth raw UWB ranging data and eliminate outliers.
Initially, the mean and variance of the raw UWB ranging data within a short temporal window are computed.
If the variance is below a preset threshold, the UWB ranging data is considered as valid and the mean is utilized to initialize the low-pass filter; otherwise, the data is discarded.
After initialization, the low-pass filter will be updated only if the latest UWB ranging is within a bound of the previous filtered UWB ranging, which is determined by the robot's maximum speed; otherwise, it remains unchanged.
The low-pass filter will be reinitialized if it has not been updated for a period of time.
Fig. \ref{figure-uwb-outlier-rejecetion} illustrates examples of raw and filtered UWB ranging data.
%Algorithm \ref{algorithm-uwb-outlier-rejection} details the UWB outlier rejection process for UWB range data between Robot $i$ and Robot $j$, and 

\subsubsection{Potential teammate detection}\label{subsubsection-potential-teammate-detection}
We employ the advanced adaptive clustering algorithm \cite{Methodology-adaptive-clustering}, a lightweight and robust point cloud clustering method, to cluster objects around the robot within the LiDAR range.
The set of these LiDAR clusters $^{b_{i}}\mathbf{C}^{t}$, generated by Robot $i$ at time $t$, may consist of surrounding obstacles as well as teammate robots. 
Each LiDAR cluster provides positional information in 
Robot $i$'s local frame, along with a 2D bounding box reflecting the size of the detected object.
However, these clusters remain anonymous to Robot $i$, lacking identity information.
To reduce the number of cluster candidates for teammate identification in subsequent modules, we pre-filter out clusters by comparing UWB ranging with the cluster position.
\begin{equation}
     ^{b_{i}}\check{\mathbf{C}}^{t} = \Big\{ {^{b_{i}}\mathbf{c}_{m}^{t}} \in {^{b_{i}}\mathbf{C}^{t}} \;\Big|\; \min_{j \in N} \Big| \big\| {^{b_{i}}\mathbf{c}_{m}^{t}} \big\| - \check{r}_{i,j}^{t} \Big| \leq 1.0 \Big\},
\end{equation}
where $\left\|. \right\|$ indicates the Euclidean distance, while $\left|. \right|$ denotes the absolute value.
The 1.0m threshold was chosen empirically to balance robustness and accuracy.
Then, we remove the clusters with large sizes from the filtered $^{b_{i}}\check{\mathbf{C}}^{t}$, as they are different from the actual robot's size.
The remaining clusters are considered as potential teammate robot clusters for subsequent processing.

\subsection{Dynamic Tracker}\label{subsection-temporary-tracker}
The remaining LiDAR clusters generated from Section \ref{subsubsection-potential-teammate-detection} do not contain ID information.
To associate clusters stemming from the same object across consecutive timestamps, we present a dynamic tracker to effectively track each cluster within the robot's own global frame.

In particular, each LiDAR cluster is assigned an independent dynamic tracker.
The state vector of the $m$-th LiDAR cluster's dynamic tracker consists of 2D position $^{G_{i}}_{c}\mathbf{p}_{m}^{t}\in \mathbb{R}^{2}$ and heading ${^{G_{i}}_{c}\theta _{m}^{t}}$.
When a new untracked LiDAR cluster $^{b_{i}}\mathbf{c}_{m}^{t}$ is generated, we allocate a new independent dynamic tracker.
This tracker is initialized by the cluster's position measurement in Robot $i$'s global frame, ${^{G_{i}}_{c}\mathbf{p}_{m}^{t}}={^{G_{i}}\mathbf{c}_{m}^{t}}={^{G_{i}}\mathbf{x}_{i}^{t}}\times {^{b_{i}}\mathbf{c}_{m}^{t}}$, and a random heading $0\leq {^{G_{i}}_{c}\theta _{m}^{t}}\leq 2\pi$.
This is because the LiDAR cluster provides only relative position measurements, without heading information.
After initialization, the dynamic tracker predicts its state vector using a random velocity model.
%Considering the motion speed of the mobile robot we used, the model has maximum linear and angular velocities of $0.2$ m/s and $0.1$ rad/s, respectively.
Subsequently, the predicted tracker is associated with the most recently generated LiDAR cluster, which has the minimum Euclidean distance to the predicted position.
For a valid association, the distance must be below a preset threshold; otherwise, the association fails.
If a valid association occurs, the dynamic tracker is corrected using the position of the corresponding LiDAR cluster, represented in Robot $i$'s global frame;
otherwise, the tracker is propagated to the next step.
The tracker will be terminated if it is propagated for too many steps without correction.
The trajectories of all tracked LiDAR clusters will be applied for teammate identification in the subsequent subsection.

\subsection{Extrinsic Transformation Estimation and Teammate Identification}\label{subsection-global-extrinsic-estimation-and-teammate-identification}
In this subsection, we aim to identify teammate robots from a set of anonymous LiDAR clusters tracked by the dynamic tracker depicted in Section \ref{subsection-temporary-tracker}.
To achieve this, we utilize a joint matching strategy, combining trajectory matching and distance matching, as described in Section \ref{subsubsection-teammate-identification}.
Before delving into the teammate identification process, it is important to note that trajectory matching is a shape-based comparison, which requires the two trajectories to be aligned for a fair comparison. In our case, the initial pose correspondence among robots is unknown, thus we have to compute the extrinsic transformation $^{G_{i}}\mathbf{T}_{j,m}^{t}\in SE(2)$, including a translation vector and a rotation matrix, to align the odometry trajectory of Robot $j$ with the trajectory of the $m$-th LiDAR cluster tracked by Robot $i$.
Moreover, this extrinsic transformation can also provide a coarse initial guess for the subsequent optimization problem in Section \ref{subsection-teammate-tracker}.

\begin{algorithm}[t]
\caption{Extrinsic transformation estimation}
\label{algorithm-global-extrinsic-transformation}
\SetAlgoVlined

\KwIn{Trajectory of the $m$-th LiDAR cluster tracked by Robot $i$, $^{G_{i}}_{c}\mathbf{p}_{m}^{t-s,t}$, and odometry trajectory shared by Robot $j$, $^{G_{j}}\mathbf{p}_{j}^{t-s,t}$.}
\KwOut{Extrinsic transformation, $^{G_{i}}\mathbf{T}_{j,m}^{t}=(\mathbf{p}, \mathbf{R})$.}

\textbf{Step 1:} compute centroids of $^{G_{i}}_{c}\mathbf{p}_{m}^{t-s,t}$ and $^{G_{j}}\mathbf{p}_{j}^{t-s,t}$. \\
\hspace{1cm} $\bar{\mathbf{p}}_{m}=\frac{1}{s}\sum_{l=t-s}^{t}{^{G_{i}}_{c}\mathbf{p}_{m}^{l}}$; \\
\hspace{1cm} $\bar{\mathbf{p}}_{j}=\frac{1}{s}\sum_{l=t-s}^{t}{^{G_{j}}\mathbf{p}_{j}^{l}}$.   \\

\textbf{Step 2:} translate the centroids to the origin. \\
\hspace{1cm} $^{G_{i}}_{c}\acute{\mathbf{p}}_{m}^{l}={^{G_{i}}_{c}\mathbf{p}_{m}^{l}}-\bar{\mathbf{p}}_{m}$;  \\
\hspace{1cm} $^{G_{j}}\acute{\mathbf{p}}_{j}^{l}={^{G_{j}}\mathbf{p}_{j}^{l}}-\bar{\mathbf{p}}_{j}$. \quad $l\in \left\{t-s, \cdots, t-1, t \right\}$.

\textbf{Step 3:} compute covariance matrix $\mathbf{H}$. \\
\hspace{1cm} $\mathbf{H}=\sum_{l=t-s}^{t}{(^{G_{i}}_{c}\acute{\mathbf{p}}_{m}^{l})^{T}}\times {^{G_{j}}\acute{\mathbf{p}}_{j}^{l}} $. \\

\textbf{Step 4:} singular value decomposition of $\mathbf{H}$. \\
\hspace{1cm} $\mathbf{H}=\mathbf{U}\mathbf{\Sigma }\mathbf{V}^{T}$. \\

\textbf{Step 5:} compute rotation matrix of $^{G_{i}}\mathbf{T}_{j,m}^{t}$. \\
\hspace{1cm} $\mathbf{R}=\mathbf{V}\mathbf{U}^{T}$.\\

\textbf{Step 6:} check reflection. \\
\hspace{1cm} \textbf{if} $\left|\mathbf{R} \right|< 0$ \textbf{then} \\
\hspace{1.5cm} $\mathbf{V}.col(1)=\mathbf{V}.col(1)\times -1$; \\
\hspace{1.5cm} $\mathbf{R}=\mathbf{V}\mathbf{U}^{T}$. \\

\textbf{Step 7:} compute translation vector of $^{G_{i}}\mathbf{T}_{j,m}^{t}$. \\
\hspace{1cm} $\mathbf{p}=\bar{\mathbf{p}}_{j}-\mathbf{R}{\bar{\mathbf{p}}_{m}}$. \\
\textbf{Reture} $^{G_{i}}\mathbf{T}_{j,m}^{t}$.
\end{algorithm}

\subsubsection{Extrinsic transformation estimation}\label{subsubsection-extrinsic-estimation}
We compute the optimal extrinsic transformation between trajectories based on the Kabsch algorithm \cite{Reference-kabsch-algorithm} by minimizing the root mean squared error.
To reduce computational cost while preserving the trajectory’s geometric properties, we retain a fixed-length window of the most recent measurements and apply downsampling to extract a set of $s+1$ evenly spaced measurements for extrinsic transformation estimation.
Algorithm \ref{algorithm-global-extrinsic-transformation} details the procedure for computing the extrinsic transformation $^{G_{i}}\mathbf{T}_{j,m}^{t}$ between odometry trajectory of Robot $j$, ${^{G_{j}}\mathbf{p}_{j}^{t-s,t}}=\left\{{^{G_{j}}\mathbf{p}_{j}^{t-s}}, \cdots, {^{G_{j}}\mathbf{p}_{j}^{t-1}}, {^{G_{j}}\mathbf{p}_{j}^{t}} \right\}$, and the trajectory of the $m$-th LiDAR cluster, ${^{G_{i}}_{c}\mathbf{p}_{m}^{t-s,t}}=\left\{{^{G_{i}}_{c}\mathbf{p}_{m}^{t-s}}, \cdots, {^{G_{i}}_{c}\mathbf{p}_{m}^{t-1}}, {^{G_{i}}_{c}\mathbf{p}_{m}^{t}} \right\}$.
It is important to note that the singular value decomposition (SVD) in Step 4 (lines 9-10) may encounter a reflection issue of the rotation matrix, causing an incorrect rotation estimation.
To avoid this issue, we assess the determinant of the obtained rotation matrix to determine whether it is a reflection.
If the determinant is negative, meaning a reflection occurs.
Then, we adjust the second column of $\mathbf{V}$ by multiplying it by $-1$, thereby ensuring that the obtained rotation matrix is correct.

\subsubsection{Teammate identification}\label{subsubsection-teammate-identification}
Ensuring correct data association between LiDAR clusters and teammate robots is crucial for accurate relative localization.
Due to the structural similarity among robots and the absence of ID markers, directly identifying nearby robots from these tracked anonymous LiDAR clusters is not feasible.
In this subsection, we introduce a joint matching strategy, combining trajectory matching and distance matching, to effectively identify teammate robots from a set of tracked anonymous LiDAR clusters.

Specifically, a tracked LiDAR cluster becomes eligible for the teammate identification process once it accumulates sufficient data.
We maintain a buffer containing the most recent measurements and uniformly sparsify it to obtain $s+1$ measurements, arranged in the form of time series.
These sparsified measurements are utilized in the subsequent matching task.
Trajectory matching and distance matching calculate trajectory dissimilarity and distance dissimilarity, based on the principle of dynamic time warping (DTW) \cite{Reference-DTW-algorithm}.
To account for variations in sequence length and ensure meaningful comparisons, the dissimilarities need to be normalized based on the length of the measurements.
For trajectory matching, the odometry trajectory of each teammate robot is compared with the trajectories of all eligible LiDAR clusters.
To guarantee fairness, each trajectory pair needs to be aligned through the latest extrinsic transformation obtained in Section \ref{subsubsection-extrinsic-estimation}.
For simplicity, we utilize the function $Trajectory\_M({^{G_{i}}\mathbf{T}_{j,m}^{t}}{^{G_{j}}\mathbf{p}_{j}^{t-s,t}}, {^{G_{i}}_{c}\mathbf{p}_{m}^{t-s,t})}$ to quantify the trajectory dissimilarity between Robot $j$'s odometry and the trajectory of the $m$-th LiDAR cluster tracked by Robot $i$, which can be calculated by
\begin{equation}
    Trajectory\_M(.)=\frac{\mathbf{D}_{t}(s,s)}{2(s+1)},
\end{equation}
where $\mathbf{D}_{t}\in \mathbb{R}^{(s+1)\times (s+1)}$ is a cost matrix, where each element represents the cumulative cost of aligning a point from ${^{G_{i}}\mathbf{T}_{j,m}^{t}}{^{G_{j}}\mathbf{p}_{j}^{t-s,t}}$ and a point from $^{G_{i}}_{c}\mathbf{p}_{m}^{t-s,t}$.
The elements of the cost matrix $\mathbf{D}_{t}$ are computed recursively as follows:
\begin{align}
    &\mathbf{D}_{t}(0,0)=\left\| {^{G_{i}}\mathbf{T}_{j,m}^{t}}{^{G_{j}}\mathbf{p}_{j}^{t}} - ^{G_{i}}_{c}\mathbf{p}_{m}^{t}\right\|, \\
    & \mathbf{D}_{t}(l,0)=\mathbf{D}_{t}(l-1,0)+\left\| {^{G_{i}}\mathbf{T}_{j,m}^{t}}{^{G_{j}}\mathbf{p}_{j}^{t-l}} - {^{G_{i}}_{c}\mathbf{p}_{m}^{t}}\right\|, \\
    & \mathbf{D}_{t}(0,h)=\mathbf{D}_{t}(0,h-1)+\left\| {^{G_{i}}\mathbf{T}_{j,m}^{t}}{^{G_{j}}\mathbf{p}_{j}^{t}} - {^{G_{i}}_{c}\mathbf{p}_{m}^{t-h}}\right\|, \\
    & \mathbf{D}_{t}(s,s)= \text{min}(\mathbf{D}_{t}(s-1,s),\mathbf{D}_{t}(s,s-1),\mathbf{D}_{t}(s-1,s-1)) \nonumber \\
    & \quad \quad \quad \quad +\left\| {^{G_{i}}\mathbf{T}_{j,m}^{t}}{^{G_{j}}\mathbf{p}_{j}^{t-s}} - {^{G_{i}}_{c}\mathbf{p}_{m}^{t-s}}\right\|,
\end{align}
where $l=1,  2, \cdots , s$ and $h=1, 2, \cdots , s$.

For distance matching, each teammate robot's UWB ranging set is compared with the distance sets of all eligible tracked LiDAR clusters.
The function $Distance\_M(\check{\mathbf{r}}_{i,j}^{t-s,t}, \mathbf{z}_{m}^{t-s,t})$ denotes the distance dissimilarity between Robot $j$'s UWB ranging set $\check{\mathbf{r}}_{i,j}^{t-s,t}=\left\{\check{r}_{i,j}^{t-s}, \cdots, \check{r}_{i,j}^{t-1}, \check{r}_{i,j}^{t} \right\}$, $i\neq j$, and the $m$-th tracked LiDAR cluster's distance set $\mathbf{z}_{m}^{t-s,t}=\left\{z_{m}^{t-s}, \cdots, z_{m}^{t-1}, z_{m}^{t} \right\}$.
Similarly, the distance dissimilarity can be computed by
\begin{equation}
    Distance\_M(.)=\frac{\mathbf{D}_{d}(s,s)}{2(s+1)}.
\end{equation}
The elements of the corresponding cost matrix $\mathbf{D}_{d}$ can be computed recursively as follows:
\begin{align}
    &\mathbf{D}_{d}(0,0)=\left | \check{r}_{i,j}^{t} - z_{m}^{t}\right |, \\
    & \mathbf{D}_{d}(l,0)=\mathbf{D}_{d}(l-1,0)+\left | \check{r}_{i,j}^{t-l} - z_{m}^{t}\right |, \\
    & \mathbf{D}_{d}(0,h)=\mathbf{D}_{d}(0,h-1)+\left | \check{r}_{i,j}^{t} - z_{m}^{t-h}\right |, \\
    & \mathbf{D}_{d}(s,s)= \text{min}(\mathbf{D}_{d}(s-1,s),\mathbf{D}_{d}(s,s-1),\mathbf{D}_{d}(s-1,s-1)) \nonumber \\
    & \quad \quad \quad \quad + \left | \check{r}_{i,j}^{t-s} - z_{m}^{t-s}\right |,
\end{align}
%where $l=1,  2, \cdots , s$ and $h=1, 2, \cdots , s$.
The $m$-th tracked LiDAR cluster's distance value $z_{m}^{t}$ can be modeled as:
\begin{equation}
    z_{m}^{t}=\left\|^{G_{i}}\mathbf{p}_{i}^{t}-{^{G_{i}}_{c}\mathbf{p }_{m}^{t}} \right\|. 
\end{equation}

To enhance the robustness and accuracy of teammate identification, we use a joint dissimilarity metric to evaluate each teammate robot and eligible cluster pair:
\begin{equation}
    {^{G_{i}}JD_{j,m}^{t}}=Trajectory\_M(.)\times Distance\_M(.),
\end{equation}
where $^{G_{i}}JD_{j,m}^{t}$ denotes a joint dissimilarity between the $m$-th LiDAR cluster tracked by Robot $i$ and teammate Robot $j$ at time $t$.
The joint dissimilarity set of all eligible clusters for Robot $j$ at time $t$ is defined as $^{G_{i}}JD_{j}^{t}$, $j\in N$ and $j\neq i$.

After that, we identify teammate Robot $j$ from these tracked anonymous LiDAR clusters according to the following rule:
\begin{equation}
    Robot\ j \leftarrow
    \left\{
    \begin{array}{ll}
        identified\ cluster, & \text{if}\ \min(^{G_{i}}JD_{j}^{t}) \leq \lambda \\ 
        not\ identified, &\text{otherwise}\\
    \end{array}
    \right.
\end{equation}
If the minimum joint dissimilarity in $^{G_{i}}JD_{j}^{t}$ is smaller than a joint dissimilarity threshold $\lambda$, the corresponding cluster is identified as Robot $j$; otherwise, not identified.
When teammate Robot $j$ is identified, its relative position in Robot $i$'s local frame can be derived from the corresponding dynamic tracker $m$, defined as $^{b_{i}}\tilde{\mathbf{p}}_{i,j}^{t}=(^{G_{i}}\mathbf{x}_{i}^{t})^{-1}\times {^{G_{i}}_{c}\mathbf{p}_{m}^{t}}$.

\begin{figure}[h]
\centering
\includegraphics[width=90mm]{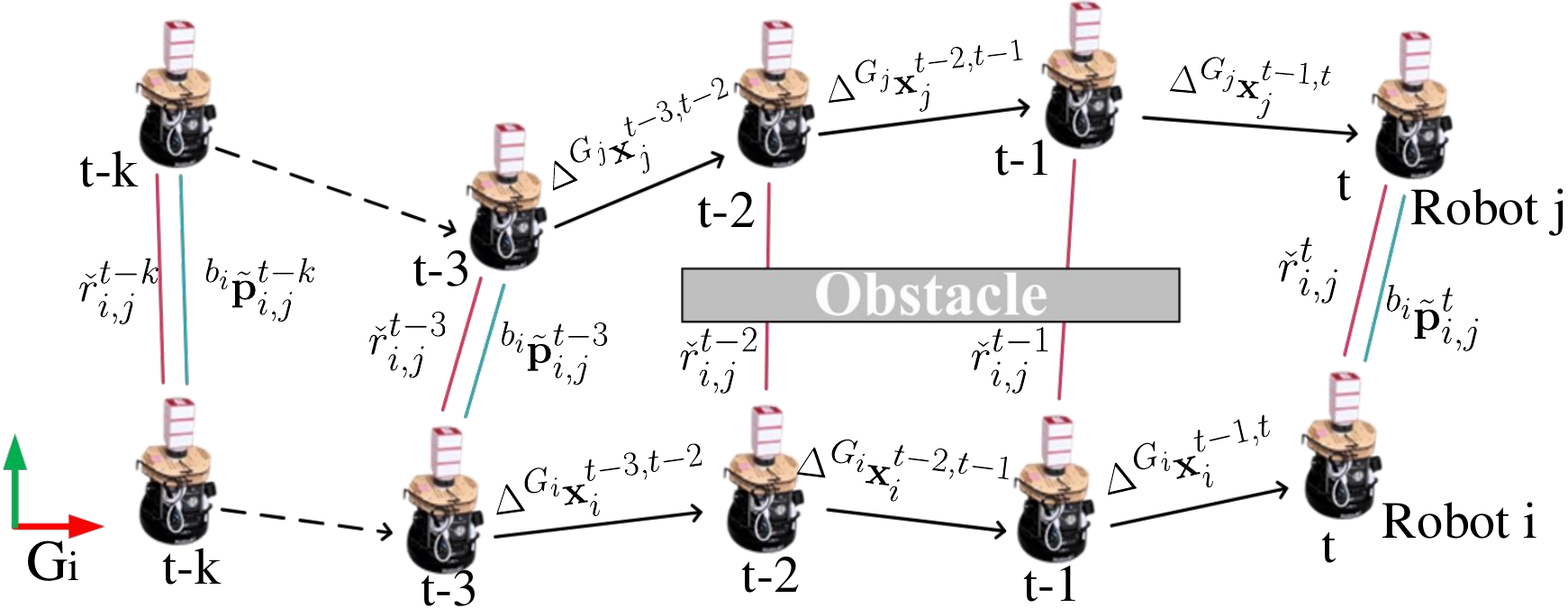}
\caption{\small Illustration of the teammate tracker based on pose graph optimization, demonstrating how Robot $i$ localizes Robot $j$ within its own global frame.
The constraints include odometry measurements of two robots (black line), UWB ranging measurements (red line), and LiDAR detection (blue line). An initial guess is used to determine the initial poses of Robot $j$'s trajectory for each optimization.}
\label{figure-teammate-tracker}
\end{figure}

\subsection{Teammate Tracker}\label{subsection-teammate-tracker}
In this subsection, we introduce the teammate tracker to track the teammate robot within the robot's own global frame, integrating UWB ranging, LiDAR, and odometry measurements.
Each teammate robot is assigned a separate teammate tracker, which becomes active when the teammate robot is identified and its cumulative displacement over a period of time exceeds a preset threshold of 2.5m, chosen empirically.
The cumulative displacement condition ensures that the rotation matrix of the extrinsic transformation estimated in Section \ref{subsubsection-extrinsic-estimation} is proper.
In this paper, we adopt a pose graph optimization-based method for the teammate tracker, as illustrated in Fig. \ref{figure-teammate-tracker}, wherein the vertex indicates the robot's pose and the edge denotes the constraints between vertices.
Given the highly nonlinear nature of the underlying optimization problem,
directly integrating the measurements without an appropriate initial guess is nearly infeasible.
To achieve effective optimization results, previous methods typically rely on repeated random sampling and select the candidate with the lowest cost as the initial guess for the optimization, like \cite{Introduction-anonymous-detection-2}\cite{Related-work-Liu}.
Instead, we provide an initial guess for each optimization by transforming the teammate robot's odometry trajectory into the robot's global frame using the latest extrinsic transformation computed from Section \ref{subsubsection-extrinsic-estimation}.
For instance, when the $m$-th cluster was identified as teammate Robot $j$, the initial guess of Robot $j$'s trajectory in Robot $i$'s global frame is set as ${^{G_{i}}\mathbf{x}_{j}^{t-k,t}}=\left\{{^{G_{i}}\mathbf{T}_{j,m}^{t}}{^{G_{j}}\mathbf{x}_{j}^{t-k}},\cdots , {^{G_{i}}\mathbf{T}_{j,m}^{t}}{^{G_{j}}\mathbf{x}_{j}^{t-1}}, {^{G_{i}}\mathbf{T}_{j,m}^{t}}{^{G_{j}}\mathbf{x}_{j}^{t}} \right\}$.
Considering the limited computational resources, we retain a fixed-length window of the most recent measurements and apply sparsification to obtain $k$ evenly spaced measurements for optimization.
The problem is formulated as follows:
\begin{equation}
\begin{aligned}
\mathop{\text{minimize}}_{{^{G_{i}}\mathbf{x}_{i}^{t-k,t}} \atop {^{G_{i}}\mathbf{x}_{j}^{t-k,t}}}\underset{odom.\; constraints\; from\; local\; Robot\; i}{\underbrace{\underset{{^{G_{i}}\mathbf{x}_{i}^{l}}\in {^{G_{i}}\mathbf{x}_{i}^{t-k+1,t}}}{\sum }\rho (\left\|{^{G_{i}}\mathbf{x}_{i}^{l}}-{^{G_{i}}\mathbf{x}_{i}^{l-1}}\times \Delta {^{G_{i}}\mathbf{x}_{i}^{l-1,l}}  \right\|_{\mathbf{p}}^{2})}} \\
+ \underset{odom.\; constraints\; from\; teammate\; Robot\; j}{\underbrace{\underset{{^{G_{i}}\mathbf{x}_{j}^{l}}\in {^{G_{i}}\mathbf{x}_{j}^{t-k+1,t}}}{\sum }\rho (\left\|{^{G_{i}}\mathbf{x}_{j}^{l}}-{^{G_{i}}\mathbf{x}_{j}^{l-1}}\times \Delta {^{G_{j}}\mathbf{x}_{j}^{l-1,l}}  \right\|_{\mathbf{p}}^{2})}} \\ 
+ \underset{LiDAR\; detection\; constraints}{\underbrace{\underset{{^{b_{i}}\tilde{\mathbf{p}}_{i,j}^{l}}\in {^{b_{i}}\tilde{\mathbf{p}}_{i,j}^{t-k,t}}}{\sum }\rho (\left\|{^{G_{i}}\mathbf{p}_{j}^{l}}-{^{G_{i}}\mathbf{p}_{i}^{l}}-{^{G_{i}}\mathbf{R}_{i}^{l}}\times {^{b_{i}}\tilde{\mathbf{p}}_{i,j}^{l}} \right\|^{2})}} \\
+ \underset{UWB\; ranging\; constraints}{\underbrace{\underset{\check{r}_{i,j}^{l}\in {\check{\mathbf{r}}}_{i,j}^{t-k,t}}{\sum }\rho (\left|\left\|{^{G_{i}}\mathbf{x}_{i}^{l}}-{^{G_{i}}\mathbf{x}_{j}^{l}} \right\|_{\mathbf{p}}- \check{r}_{i,j}^{l}\right|^{2})}}\ ,
\end{aligned}
\end{equation}
where $^{b_{i}}\tilde{\mathbf{p}}_{i,j}^{t-k,t}$ represents the set of LiDAR detection constraints from time $t-k$ to $t$ obtained from the teammate identification module in Section \ref{subsubsection-teammate-identification}.
Note that although the LiDAR-based constraint may be unavailable under occlusion, the teammate tracker can still operate reliably using only odometry and UWB ranging measurements.
$\left\|. \right\|_{\mathbf{p}}$ denotes the Euclidean distance of the translation component.
$\Delta(.)$ indicates the displacement between consecutive timestamps, measured by odometry.
To reduce the impact caused by the possible outliers in constraints, Huber norm $\rho (.)$ is utilized, with thresholds set empirically to 0.025m, 0.5m, and 1.0m for odometry, LiDAR, and UWB ranging constraints, respectively.
The g2o optimization library \cite{Reference-g2o} is adopted to solve Eq. 15.
The corresponding information matrices are computed as the inverse of covariance matrices, which are empirically determined based on fixed standard deviations: 0.005m and 0.005rad for odometry, 0.1m for LiDAR detection, and 0.2m for UWB ranging.
Then, we can obtain the pose of Robot $j$ in Robot $i$'s global frame at time $t$.
Afterward, the relative pose of Robot $j$ with respect to Robot $i$ at time $t$ can be calculated by a coordinate transformation, $^{b_{i}}\mathbf{x}_{i,j}^{t}=({^{G_{i}}\mathbf{x}_{i}^{t}})^{-1}\times {^{G_{i}}\mathbf{x}_{j}^{t}}$.
A point worth noting is that the inter-robot constraints lack the heading information, owing to the inherent limitations of LiDAR and UWB ranging measurements.
%Note that each teammate robot is assigned an individual teammate tracker, ensuring real-time performance.

\subsection{UWB Communication Network}\label{subsection-uwb-communication-network}
\begin{figure}[t]
\centering
\includegraphics[width=85mm]{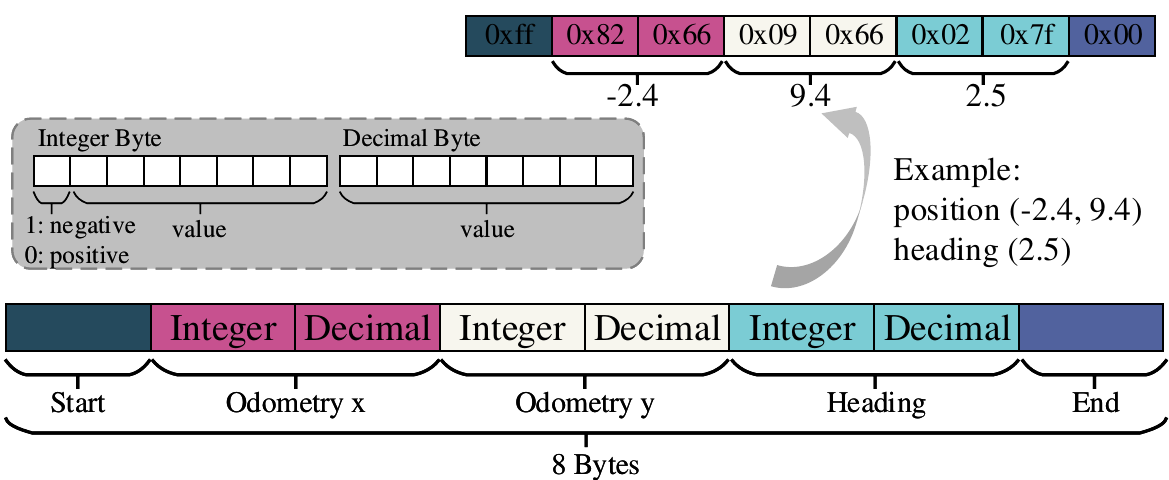}
\caption{\small The message structure employed in the UWB communication network. Only odometry information is broadcasted within our multi-robot system.}
\label{figure-uwb-communication-network}
\end{figure}

Each robot only broadcasts its odometry measurement to others solely based on a UWB communication network, without the need for any external communication devices, such as WiFi router.
The UWB sensor utilized in our system supports Byte-level data transmission when performing ranging, with a maximum payload of approximately 100 Bytes.
However, this limited bandwidth is insufficient for transmitting high-dimensional information, such as point clouds or feature descriptors.
To minimize communication bandwidth usage while ensuring reliability and accuracy, we represent each odometry measurement using only 8 Bytes.
Fig. \ref{figure-uwb-communication-network} illustrates the message structure employed in the UWB communication network.
Particularly, the first and last Byte serve as starting and ending markers, respectively.
Since our system is implemented in a 2D environment, only 2D position and heading information are transmitted.
Each value is encoded using two consecutive Bytes: one for the integer part and the other for the decimal part, with a resolution of approximately 0.004.
The transmissible position range spans from -127.996m to 127.996m.
The heading is expressed in radian to cover the full range from $0$ to $2\pi$.
This lightweight data exchange is fundamental to sustaining reliable relative pose estimation.
%To the best of our knowledge, this is the first multi-robot system that relies solely on the UWB communication network, without external communication devices.

\section{Experimental Results}\label{section-experiments}

\subsection{Experimental Setup}\label{subsection-experimental-setup}
\begin{figure}[t]
\centering
\includegraphics[width=82mm]{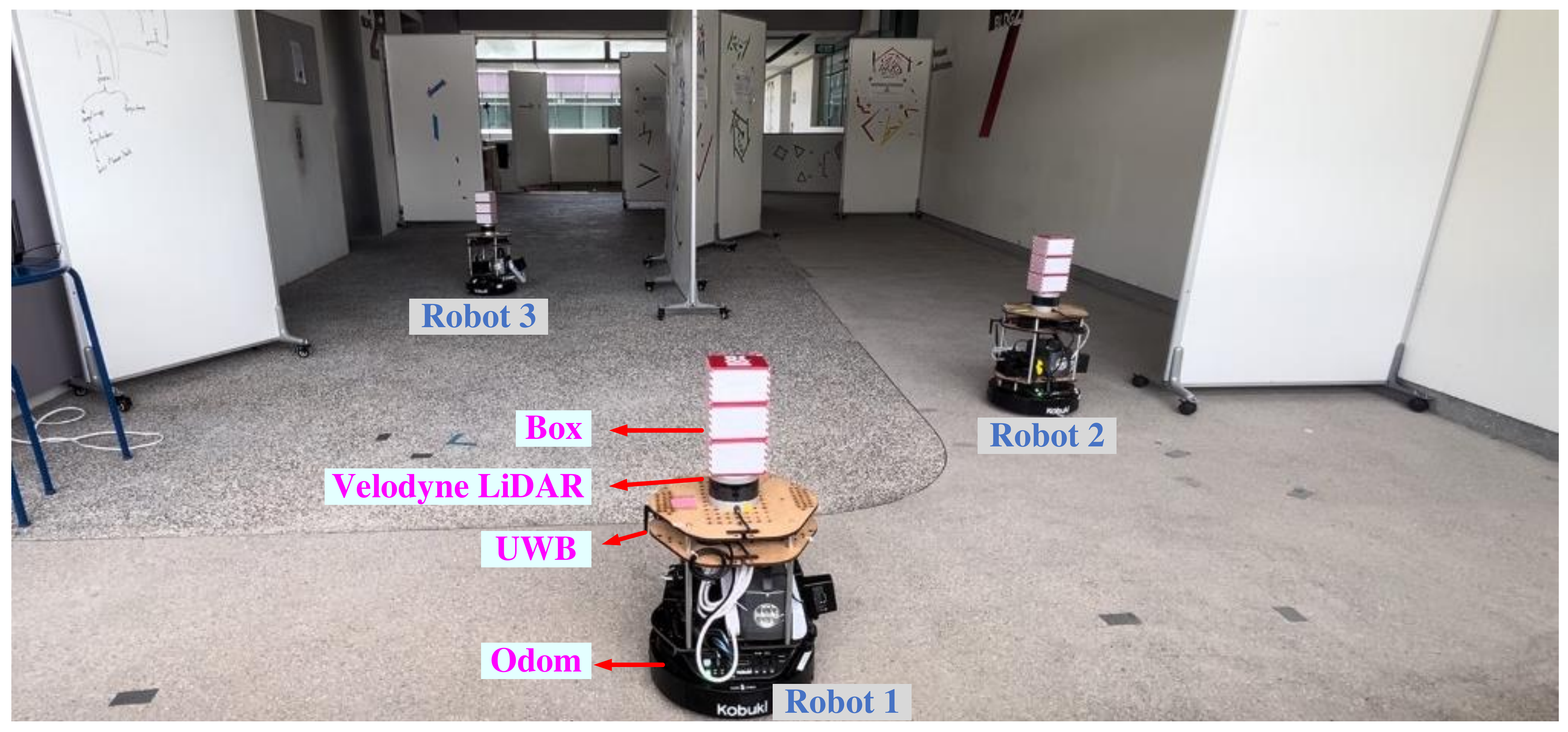}
\caption{\small Three mobile robots moved randomly in the environment with a size of 17.1m$\times$6.2m, and the initial correspondence among them is unknown. The box is used for controllable LiDAR detection, ensuring a more systematic performance evaluation. Notably, only a single LiDAR scan channel is used for 2D perception.}
\label{figure-sectioniv-experiment-environment}
\end{figure}

To validate the performance of the proposed approach, we have conducted experiments in an indoor environment with three homogeneous TurtleBot2 robots.
As shown in Fig. \ref{figure-sectioniv-experiment-environment}, each robot was equipped with one UWB node (Nooploop LinkTrack, maximum range of 100m, ranging frequency of 50Hz, maximum data transmission of 100 Bytes at frequency of 20Hz) and LiDAR (Velodyne VLP-16, maximum range of 100m with horizontal field of view of 360$^{\circ}$ at frequency of 10Hz).
Notably, since this study focuses on 2D scenarios, only a single LiDAR scan channel is used in the experiment, which aligns with the 2D nature of the state estimation and the wheel odometry employed in our system.
Each robot was controlled to randomly move at a maximum speed of 0.2m/s and provide odometry measurements with a sampling frequency of 50Hz.
The proposed approach is evaluated on a computing unit with an E5-2620 v4 CPU, running Robot Operating System (ROS) Noetic.
During the experiments, the initial correspondence among robots is undisclosed, and odometry measurement is broadcast via the UWB communication network depicted in Section \ref{subsection-uwb-communication-network}.
Adaptive Monte Carlo Localization\footurl{https://wiki.ros.org/amcl}{AMCL} was applied to each robot to obtain the ground truth based on a prior occupancy grid map.
%To guarantee real-time performance, the proposed system leverages several independent threads implemented in C++.

\begin{figure}[t]
\centering
\includegraphics[width=88mm]{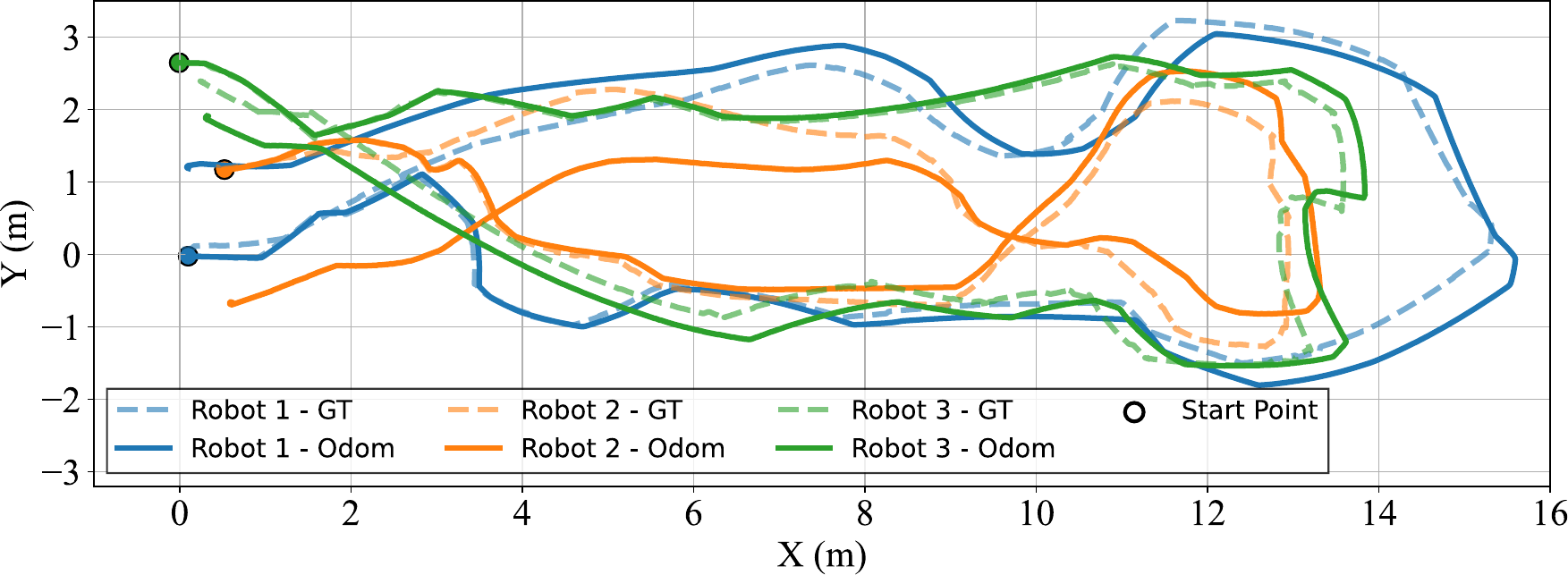}
\caption{\small Ground truth and odometry measurements of three different robots in the experiment.}
\label{figure-gt-odom}
\end{figure}

For visualization and comparison purposes, all trajectories were manually transformed into a common reference frame in the offline processing stage, as illustrated in Fig. \ref{figure-gt-odom}. Each robot's odometry measurements experience considerable drift.

\subsection{Evaluation of Teammate Identification}\label{subsection-results-of-teammate-identification}
\begin{figure}[t]
    \vspace{-\baselineskip}
    \centering
    \includegraphics[width=80mm]{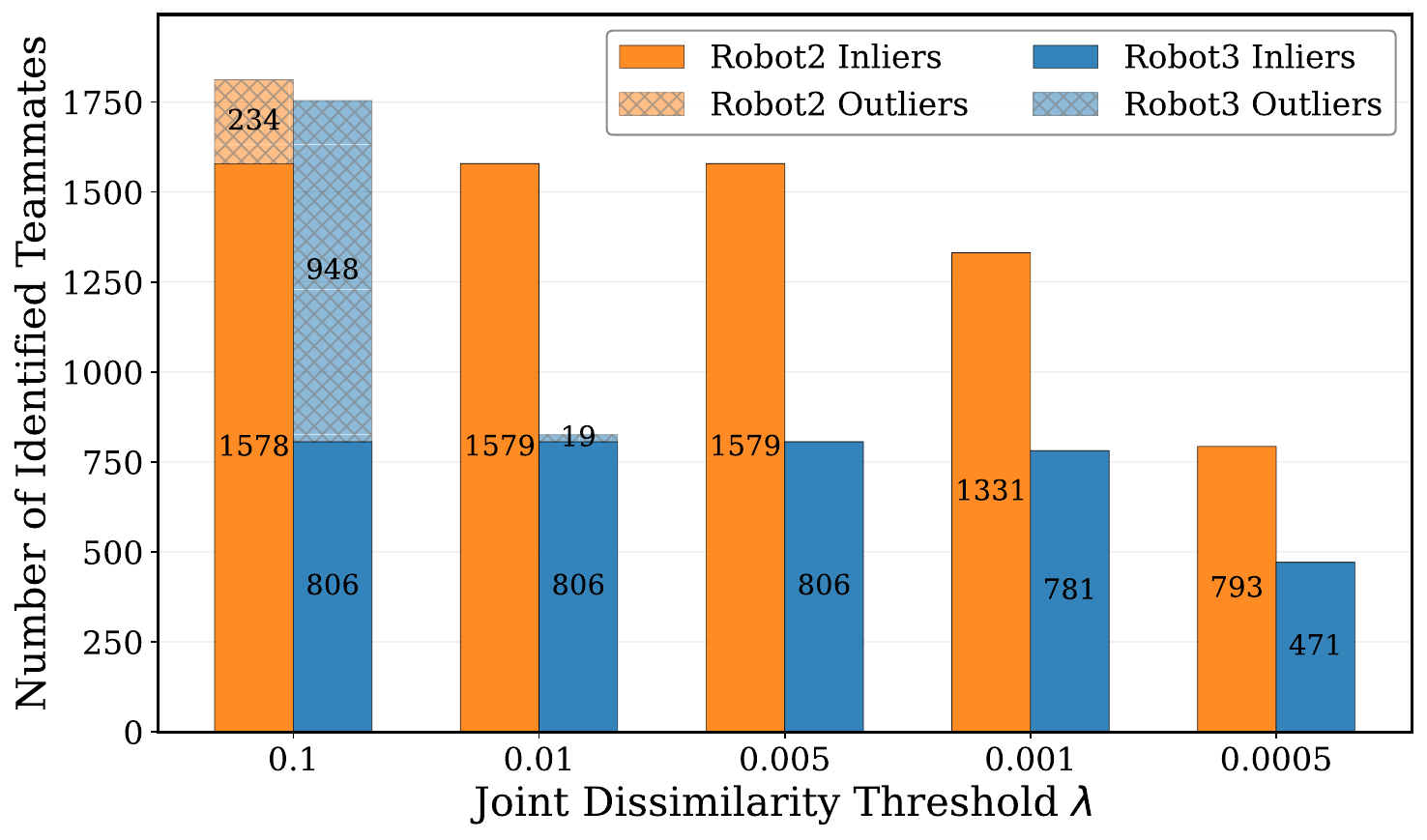}
    \caption{\small Evaluation of teammate identification using various joint dissimilarity thresholds $\lambda$.}
    \label{figure-joint-dissimilarity-threshold}
\end{figure}

Accurately identifying teammate robots from the anonymous LiDAR clusters is critical to the proposed system, as it directly influences the performance of subsequent modules.
We define those erroneous associations as outliers, while the correct associations are regarded as inliers.
%We fixed the number of measurements $s$ to 50.
We varied the joint dissimilarity threshold $\lambda$ from 0.1 to 0.0005 to examine the performance of the proposed teammate identification method.
As shown in Fig. \ref{figure-joint-dissimilarity-threshold}, a higher joint dissimilarity threshold $\lambda$ results in more associations; however, the proportion of outliers is also substantial, which reduces the identification accuracy.
When the threshold $\lambda$ reduces to 0.005, the outliers are fully excluded, while the number of inliers remains virtually unaffected.
As $\lambda$ is gradually reduced, although the stricter criterion effectively eliminates outliers, the number of inliers also decreases, indicating that some valid associations are also discarded along with erroneous associations.

Additionally, we evaluate teammate identification using different methods, namely pure distance matching, pure trajectory matching, and joint matching (i.e., distance matching + trajectory matching), as depicted in Section \ref{subsubsection-teammate-identification}.
We set the distance and trajectory dissimilarity threshold to 0.3 and fixed the joint dissimilarity threshold $\lambda$ to 0.005.
Table \ref{table-teammate-identification-evaluation} summarizes the experimental results of teammate identification under different matching methods from the perspective of Robot 1.
As shown in this table, the pure distance matching-based approach produces a lot of outliers for Robot 2 and Robot 3.
In pure trajectory matching, Robot 3 is particularly prone to outliers because its odometry trajectory is less distinctive compared to those of the other LiDAR cluster objects.
Note that distance matching does not require communication, whereas trajectory matching relies on the shared odometry measurements from other teammate robots.
In contrast, the joint matching strategy (distance matching + trajectory matching) outperforms the individual matching methods, effectively avoiding false identifications.
The average relative translational error for Robot 2 and Robot 3 is 0.08m and 0.13m, respectively.
However, it is worth noting that the LiDAR cluster of the teammate robot may not always be available due to occlusion.
Moreover, the random movement of each robot created variable occlusion conditions, resulting in differences in the number of inliers.

\begin{table*}[t]
\centering
\setlength{\abovecaptionskip}{0pt}%    
\setlength{\belowcaptionskip}{0pt}%
\caption{\small Evaluation of teammate identification under different teammate identification methods from Robot 1's perspective: number of identified teammates and translational error (in meters).}
\label{table-teammate-identification-evaluation}
\begin{tabular}{|c|ccc|ccc|}
\hline
\multirow{2}{*}{Approach}                                                        & \multicolumn{3}{c|}{Robot 2}                                                                  & \multicolumn{3}{c|}{Robot 3}                                                                  \\ \cline{2-7} 
                                                                                 & \multicolumn{1}{c|}{No. of Inliers} & \multicolumn{1}{c|}{No. of Outliers} & Trans. error (m) & \multicolumn{1}{c|}{No. of Inliers} & \multicolumn{1}{c|}{No. of Outliers} & Trans. error (m) \\ \hline
Distance matching                                                                & \multicolumn{1}{c|}{1579}           & \multicolumn{1}{c|}{80}              & 0.15            & \multicolumn{1}{c|}{806}            & \multicolumn{1}{c|}{259}             & 1.08            \\ \hline
Trajectory matching                                                              & \multicolumn{1}{c|}{1579}           & \multicolumn{1}{c|}{0}               & 0.08            & \multicolumn{1}{c|}{806}            & \multicolumn{1}{c|}{279}             & 1.01            \\ \hline
\begin{tabular}[c]{@{}c@{}}Joint matching (\textbf{Ours})\end{tabular} & \multicolumn{1}{c|}{1579}           & \multicolumn{1}{c|}{0}               & 0.08            & \multicolumn{1}{c|}{806}            & \multicolumn{1}{c|}{0}               & 0.13            \\ \hline
\end{tabular}
\end{table*}

\subsection{Evaluation of Teammate Tracker}\label{subsection-results-of-teammate-tracker}
\begin{figure}[t]
\centering
\vspace{-\baselineskip}
\setlength{\abovecaptionskip}{0pt}    
\setlength{\belowcaptionskip}{6pt}
\subfloat[Estimated trajectory of Robot 2]{
    \includegraphics[width=85mm]{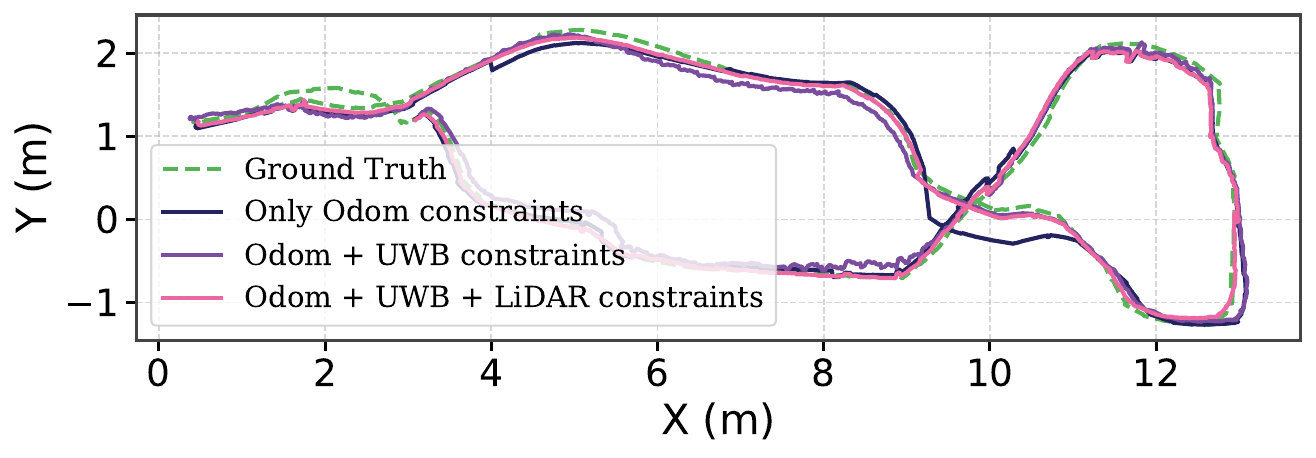}
}

\subfloat[Estimated trajectory of Robot 3]{
    \includegraphics[width=85mm]{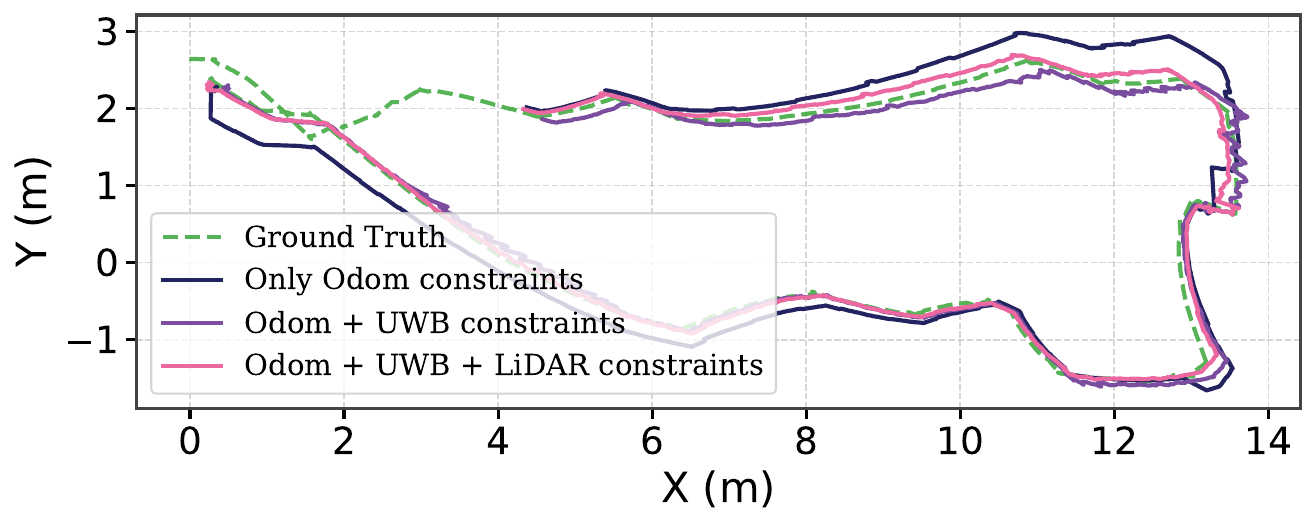}
}
\caption{\small The estimated trajectory of Robot 2 and Robot 3 with respect to Robot 1, generated by the teammate tracker with an initial guess based on the latest extrinsic transformation for each optimization under different combinations of constraints (dark blue line: odom; purple line: odom+UWB; pink line: odom+UWB+LiDAR).}
\label{figure-estimated-trajectory}
\end{figure}

In this subsection, we evaluate the performance of the teammate tracker in Section \ref{subsection-teammate-tracker}.
In our experiment, Robot 1 and Robot 3 moved first, followed by Robot 2 after around 40 seconds.
Fig. \ref{figure-estimated-trajectory} illustrates the estimated trajectory of Robot 2 and Robot 3 with respect to Robot 1, produced by the teammate tracker with different constraint conditions.
For trajectory visualization, the estimated trajectories of Robot 2 and Robot 3 are obtained by adding the estimated relative poses to the ground truth trajectory of the observer Robot 1.
As observed from Fig. \ref{figure-estimated-trajectory}, when only odometry constraint is available, the performance of the teammate tracker might degrade over time; however, it will recover after a while.
This recovery occurs as the corresponding extrinsic transformation is periodically updated, providing a better initial guess for the optimization.
The accuracy is improved with the integration of UWB and LiDAR constraints, as shown in Fig. \ref{figure-estimated-trajectory}.
Table \ref{table-teammate-tracker-computational-time} lists the average time consumption per operation for the teammate tracker, as well as the relative localization accuracy under varying constraint conditions.
Computational time tends to increase as the constraint conditions become more informative.
Each teammate robot is localized using an independent teammate tracker, with all trackers executed in parallel to ensure scalability and computational efficiency. As a result, the average time consumption per teammate tracker is approximately 50 ms.
In addition, it is worthy of note that the initial estimations for Robot 2 and Robot 3 occurred at different moments.  
This is because each teammate robot began moving at different timestamps and required sufficient cumulated displacement to trigger the respective teammate tracker.
This point can be found in our attached video.
After that, the teammate robots could be continuously estimated, even in the presence of occlusions.
%Note that each teammate tracker is executed in a separate thread to ensure real-time performance.

\begin{table*}[t]
\centering
\setlength{\abovecaptionskip}{0pt}%    
\setlength{\belowcaptionskip}{0pt}%
\caption{\small Evaluation of the teammate tracker with different constraint conditions from Robot 1's perspective: relative translational error (in meters), relative rotational error (in degrees), and average computational time (in milliseconds). mean is the average value, and std represents the standard deviation value.}
\label{table-teammate-tracker-computational-time}
\begin{tabular}{|c|ccccc|ccccc|ccccc|}
\hline
\multirow{3}{*}{} & \multicolumn{5}{c|}{Odom constraints}                                                                                                                                                              & \multicolumn{5}{c|}{Odom + UWB constraints}                                                                                                                                                        & \multicolumn{5}{c|}{\textbf{Odom + UWB + LiDAR constraints}}                                                                                                                                                \\ \cline{2-16} 
                  & \multicolumn{2}{c|}{Trans. error}                         & \multicolumn{2}{c|}{Rot. error}                & \multirow{2}{*}{\begin{tabular}[c]{@{}c@{}}Time\\ (ms)\end{tabular}} & \multicolumn{2}{c|}{Trans. error}                         & \multicolumn{2}{c|}{Rot. error}                & \multirow{2}{*}{\begin{tabular}[c]{@{}c@{}}Time\\ (ms)\end{tabular}} & \multicolumn{2}{c|}{Trans. error}                         & \multicolumn{2}{c|}{Rot. error}                & \multirow{2}{*}{\begin{tabular}[c]{@{}c@{}}Time\\ (ms)\end{tabular}} \\ \cline{2-5} \cline{7-10} \cline{12-15}
                  & \multicolumn{1}{c|}{mean}  & \multicolumn{1}{c|}{std}   & \multicolumn{1}{c|}{mean} & \multicolumn{1}{c|}{std}  &                                                                      & \multicolumn{1}{c|}{mean}  & \multicolumn{1}{c|}{std}   & \multicolumn{1}{c|}{mean} & \multicolumn{1}{c|}{std}  &                                                                      & \multicolumn{1}{c|}{mean}  & \multicolumn{1}{c|}{std}   & \multicolumn{1}{c|}{mean} & \multicolumn{1}{c|}{std}  &                                                                      \\ \hline
Robot 2           & \multicolumn{1}{c|}{0.15} & \multicolumn{1}{c|}{0.13} & \multicolumn{1}{c|}{3.82} & \multicolumn{1}{c|}{2.73} & 33.1                                                                 & \multicolumn{1}{c|}{0.11} & \multicolumn{1}{c|}{0.07} & \multicolumn{1}{c|}{3.76} & \multicolumn{1}{c|}{2.22} & 49.5                                                                 & \multicolumn{1}{c|}{0.09} & \multicolumn{1}{c|}{0.05} & \multicolumn{1}{c|}{2.95} & \multicolumn{1}{c|}{1.39} & 50.5                                                                 \\ \hline
Robot 3           & \multicolumn{1}{c|}{0.31} & \multicolumn{1}{c|}{0.18} & \multicolumn{1}{c|}{4.72} & \multicolumn{1}{c|}{2.29} & 34.3                                                                 & \multicolumn{1}{c|}{0.21} & \multicolumn{1}{c|}{0.13} & \multicolumn{1}{c|}{3.34} & \multicolumn{1}{c|}{1.99} & 50.4                                                                 & \multicolumn{1}{c|}{0.14} & \multicolumn{1}{c|}{0.06} & \multicolumn{1}{c|}{3.20} & \multicolumn{1}{c|}{1.50} & 51.1                                                                 \\ \hline
\end{tabular}
\end{table*}

\begin{figure}[t]
\centering
\vspace{-\baselineskip}
\setlength{\abovecaptionskip}{0pt}    
\setlength{\belowcaptionskip}{0pt}
\begin{tabular}{cc}
    \includegraphics[width=0.23\textwidth]{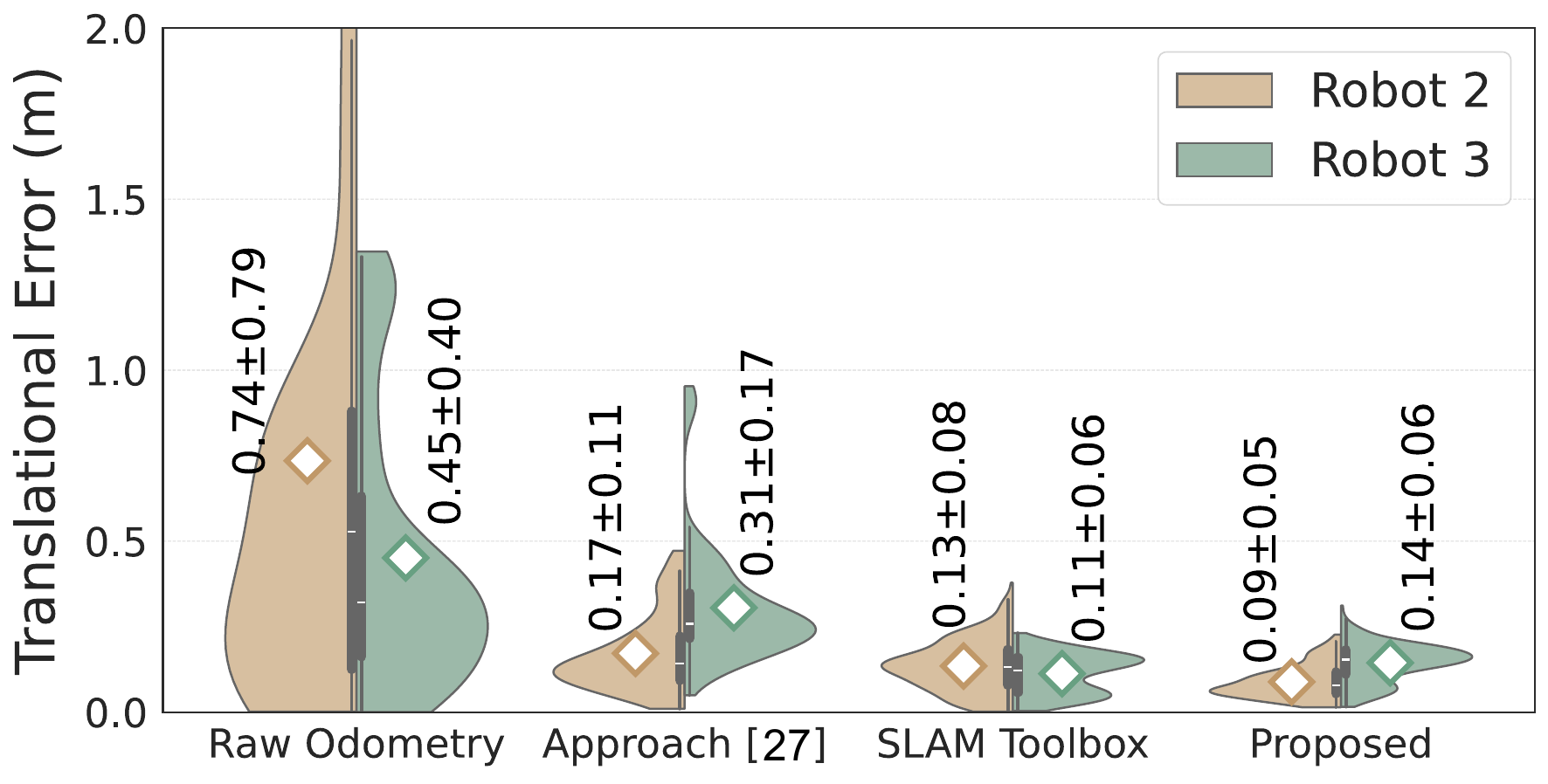} &
    \includegraphics[width=0.23\textwidth]{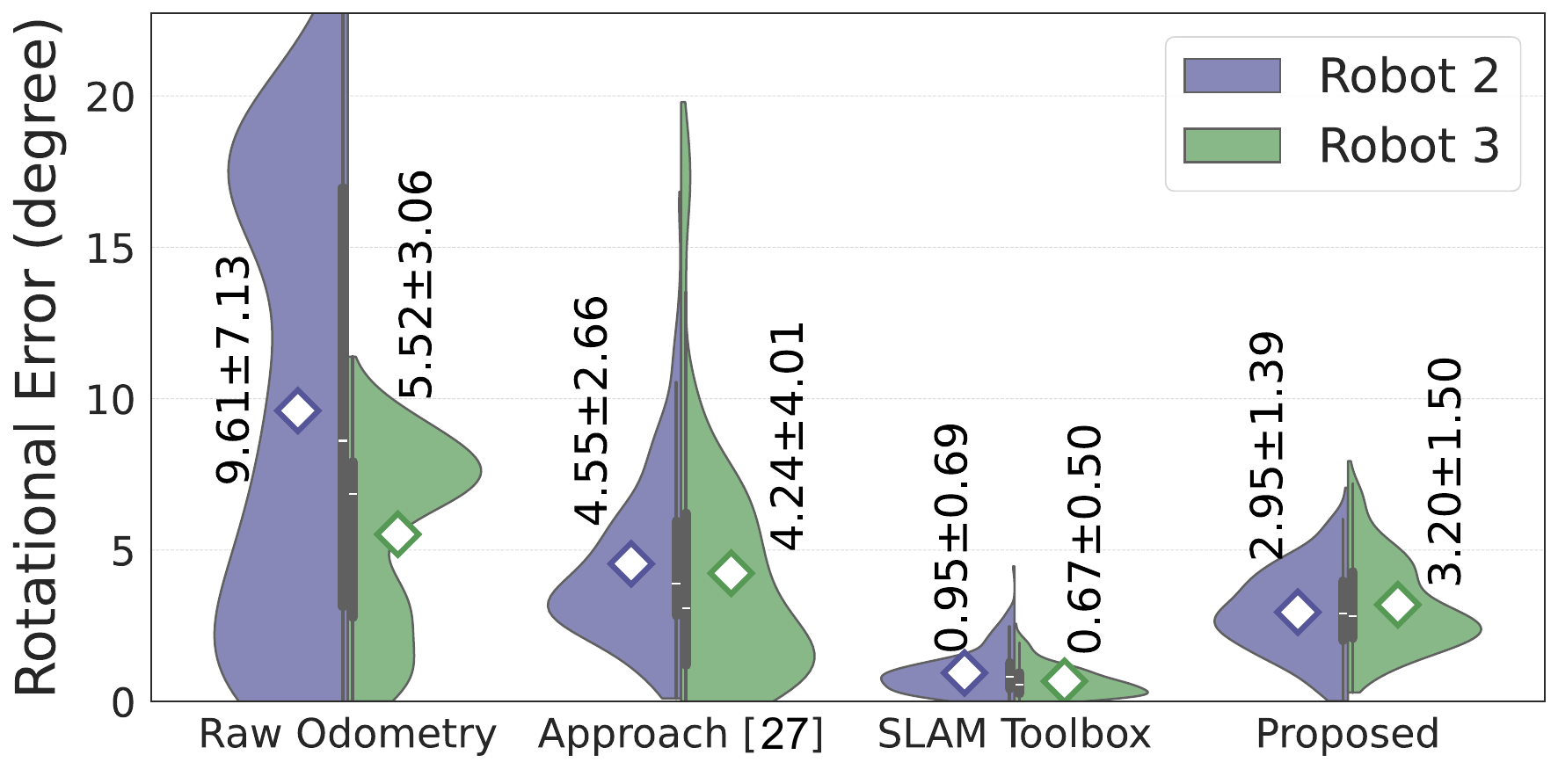} \\
    \multicolumn{2}{c}{\small (a) Robot 1 localizes others} \\[0.5em]
    
    \includegraphics[width=0.23\textwidth]{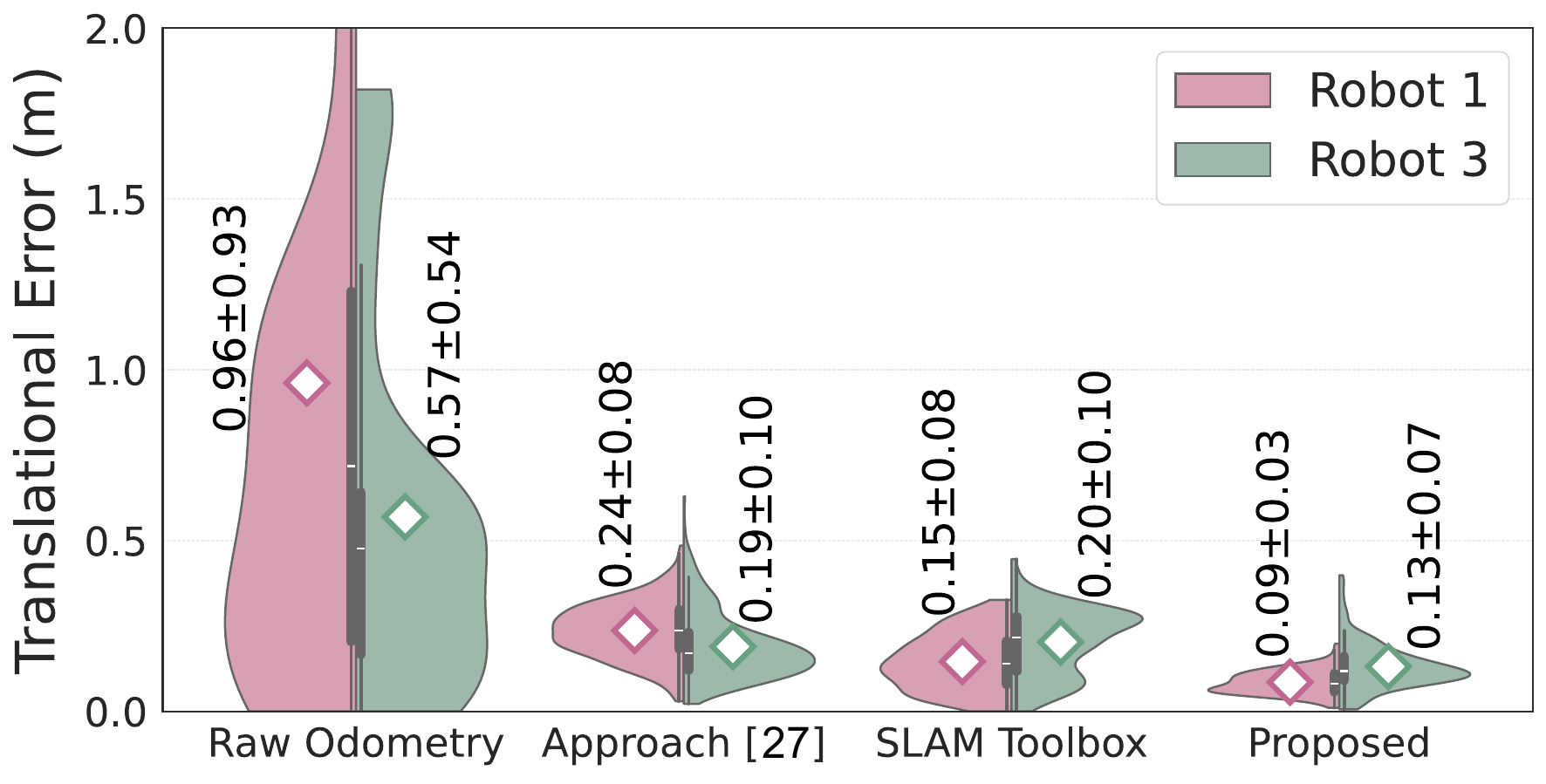} &
    \includegraphics[width=0.23\textwidth]{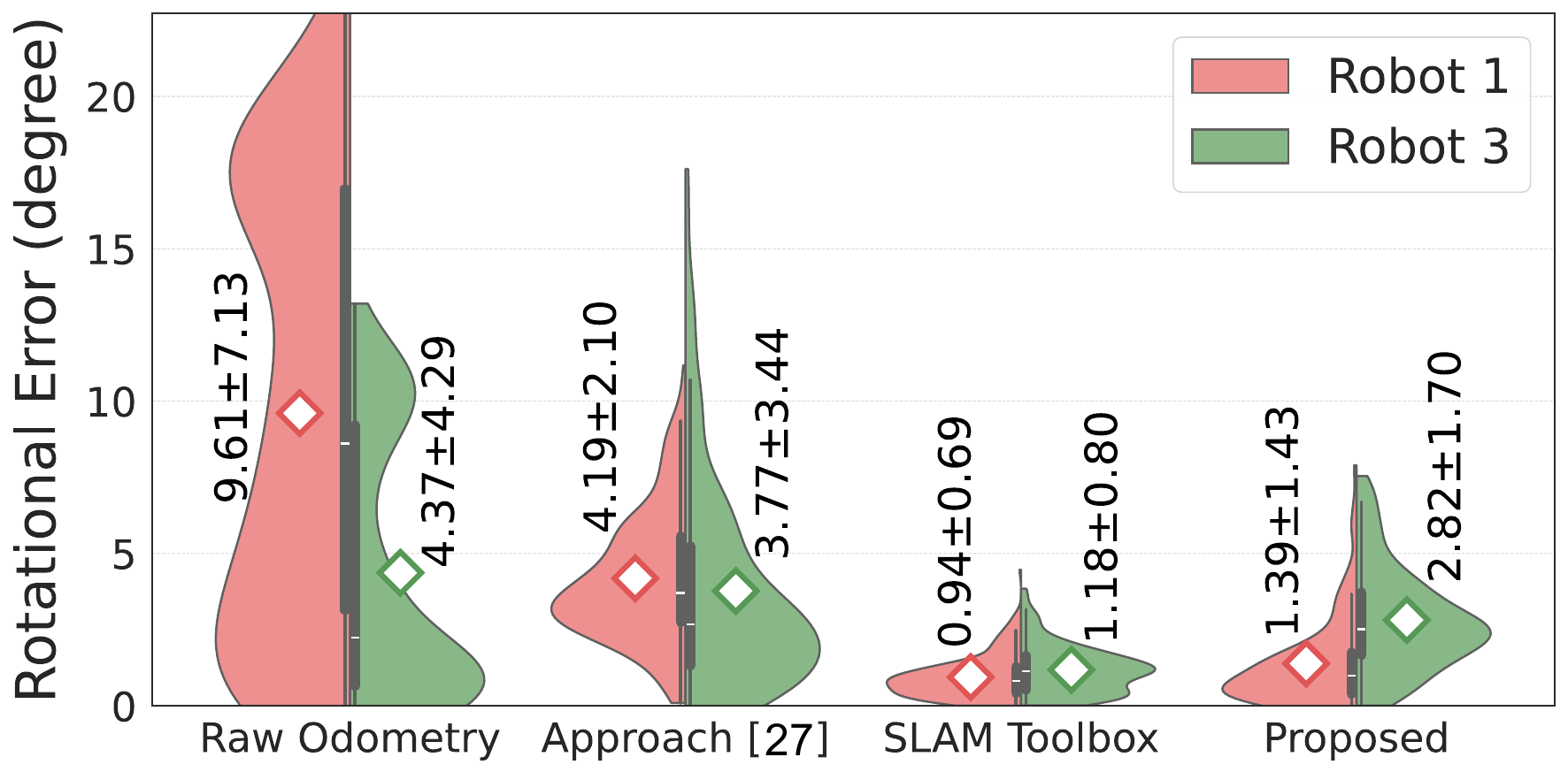} \\
    \multicolumn{2}{c}{\small (b) Robot 2 localizes others} \\[0.5em]
    
    \includegraphics[width=0.23\textwidth]{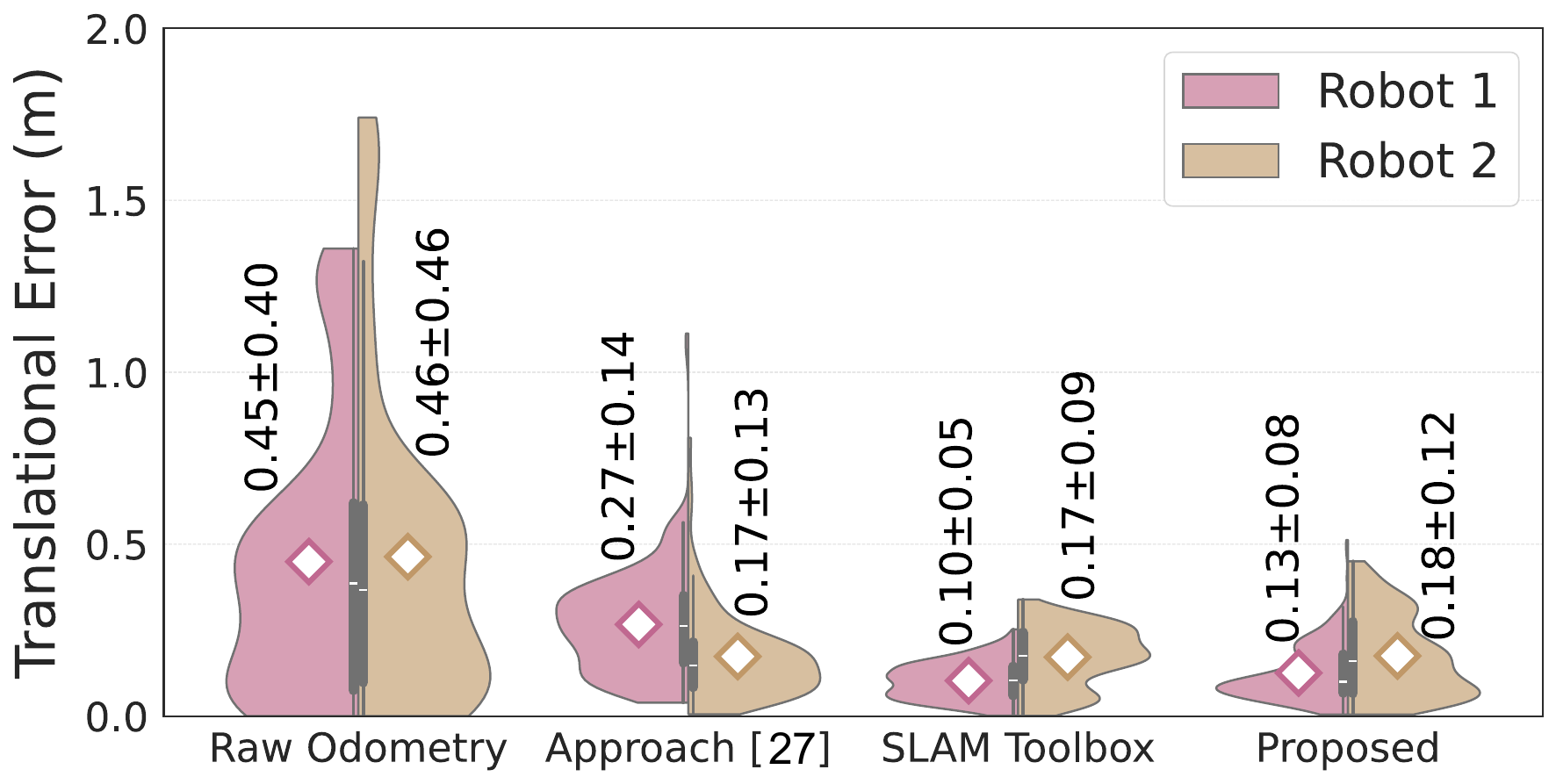} &
    \includegraphics[width=0.23\textwidth]{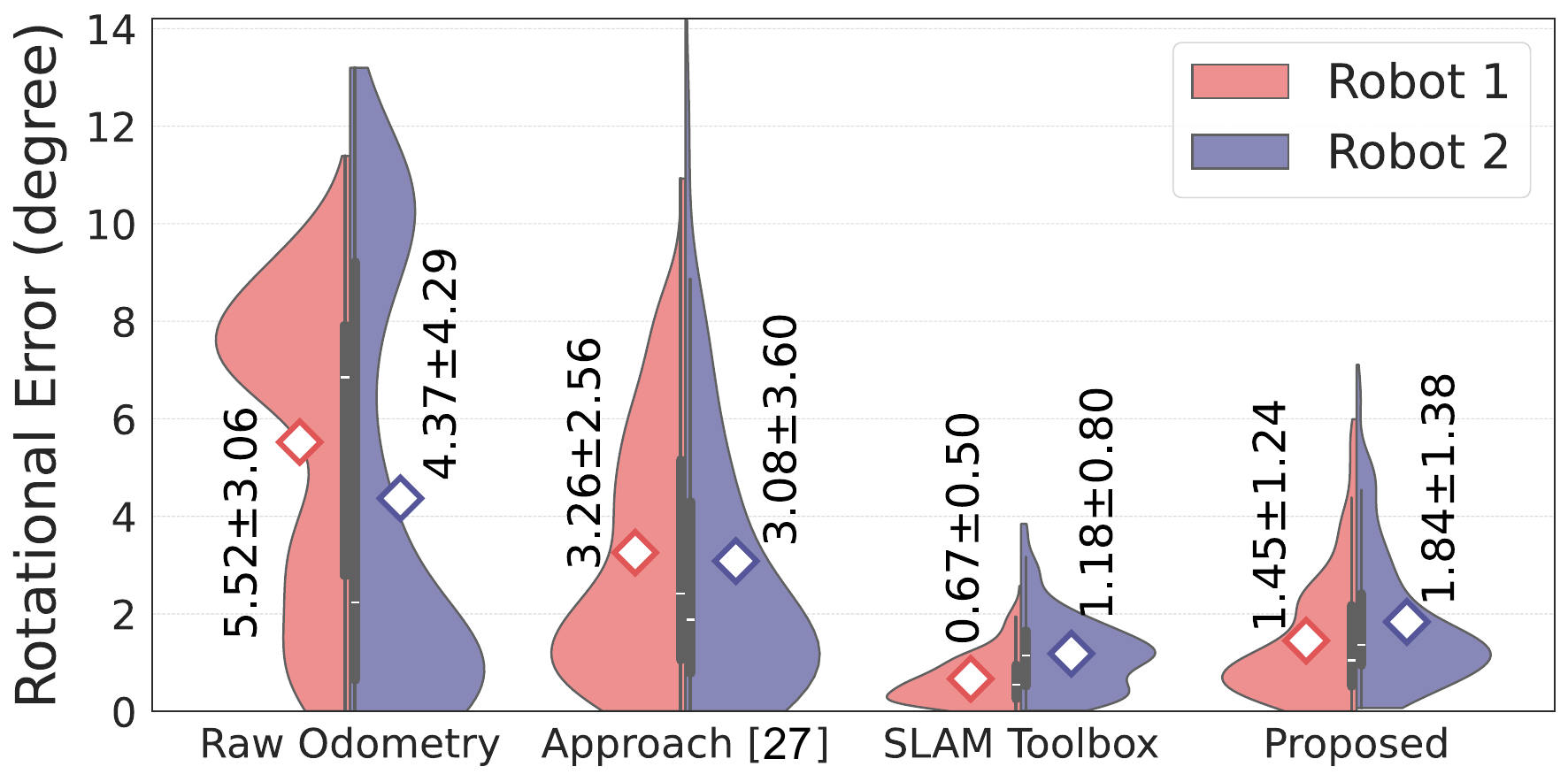} \\
    \multicolumn{2}{c}{\small (c) Robot 3 localizes others} \\
\end{tabular}
\caption{\small Evaluation of different approaches with identical communication requirements: the average relative translational error (in meters) and relative rotational error (in degrees) in our experiment.}
\label{figure-comparison}
\end{figure}

\subsection{Comparison with Existing Approach}\label{subsection-experiment-comparison}
We evaluate the proposed approach against two representative methods.
The first baseline \cite{Related-work-Liu} estimates relative poses by combining short-term UWB ranging and odometry, without requiring initial pose correspondence among robots.
The second is SLAM Toolbox \cite{slam-toolbox}, a LiDAR–odometry SLAM framework, where relative poses can be derived by exchanging self-localization results, under the assumption of known initial pose correspondence.
All baselines incur communication costs similar to our approach.
Fig. \ref{figure-comparison} demonstrates a comparison of the relative translational error and relative rotational error across the different approaches, in which the relative localization is conducted from each robot's perspective.
As can be observed from this figure, our approach achieves the average translational error of 0.126m and rotational error of 2.275$^{\circ}$.
Compared with all methods, our approach attains the highest translational accuracy and exhibits a low standard deviation, indicating strong robustness and reliability.
In terms of rotational accuracy, the proposed method significantly outperforms pure odometry and the approach in \cite{Related-work-Liu}.
In contrast, SLAM Toolbox \cite{slam-toolbox} benefits from LiDAR’s strong angular observability, which enables more reliable orientation corrections and thus yields higher rotational accuracy than our approach.
This outcome is expected, as the proposed approach lacks explicit orientation constraints.

\begin{table}[t]
\centering
\vspace{-\baselineskip}
\setlength{\abovecaptionskip}{0pt}%    
\setlength{\belowcaptionskip}{0pt}%
\caption{\small Packet loss rate in our experiment using the UWB communication network. One packet corresponds to one message consisting of 8 Bytes.}
\label{table-packet-loss-rate}
\resizebox{0.49\textwidth}{!}{
\begin{tabular}{|c|cc|cc|c|}
\hline
\multirow{2}{*}{} & \multicolumn{2}{c|}{What should be received}   & \multicolumn{2}{c|}{What has received}         & \multirow{2}{*}{\begin{tabular}[c]{@{}c@{}}Loss rate\\ (\%)\end{tabular}} \\ \cline{2-5}
                  & \multicolumn{1}{c|}{\# of packets} & Size (KB) & \multicolumn{1}{c|}{\# of packets} & Size (KB) &                                                                           \\ \hline
Robot 1           & \multicolumn{1}{c|}{9597}          & 76.776    & \multicolumn{1}{c|}{9536}          & 76.288    & 0.64\%                                                                    \\ \hline
Robot 2           & \multicolumn{1}{c|}{9595}          & 76.760    & \multicolumn{1}{c|}{9564}          & 76.512    & 0.32\%                                                                    \\ \hline
Robot 3           & \multicolumn{1}{c|}{9596}          & 76.768    & \multicolumn{1}{c|}{9551}          & 76.408    & 0.47\%                                                                    \\ \hline
\end{tabular}}
\end{table}

\subsection{Evaluation of Communication Efficiency}\label{subsection-experiment-communication}
In the proposed system, broadcasting odometry measurement only relies on the UWB communication network, without external communication devices.
In this experiment with three robots, the total communication bandwidth required is only 0.48KB/s at a communication frequency of 20Hz, i.e., message size (8 Bytes) $\times$ broadcast frequency (20Hz) $\times$ number of robots (3), resulting in a total data exchange of 115.2KB for 240s duration of our experiment.
Table \ref{table-packet-loss-rate} reports the packet loss rate for each robot in our experiment, using the UWB communication network presented in Section \ref{subsection-uwb-communication-network}.
The results show that the average packet loss rate is around 0.476\% across three robots during this experiment.
The proposed approach enables relative localization for a group of robots with instantaneous UWB ranging and communication, which is well-suited for the coordination and formation of miniaturized vehicles with limited onboard resources.

\begin{figure}[h]
    \centering
    \vspace{-\baselineskip}
    \setlength{\abovecaptionskip}{0pt}%
    \setlength{\belowcaptionskip}{0pt}%
    \includegraphics[width=86mm]{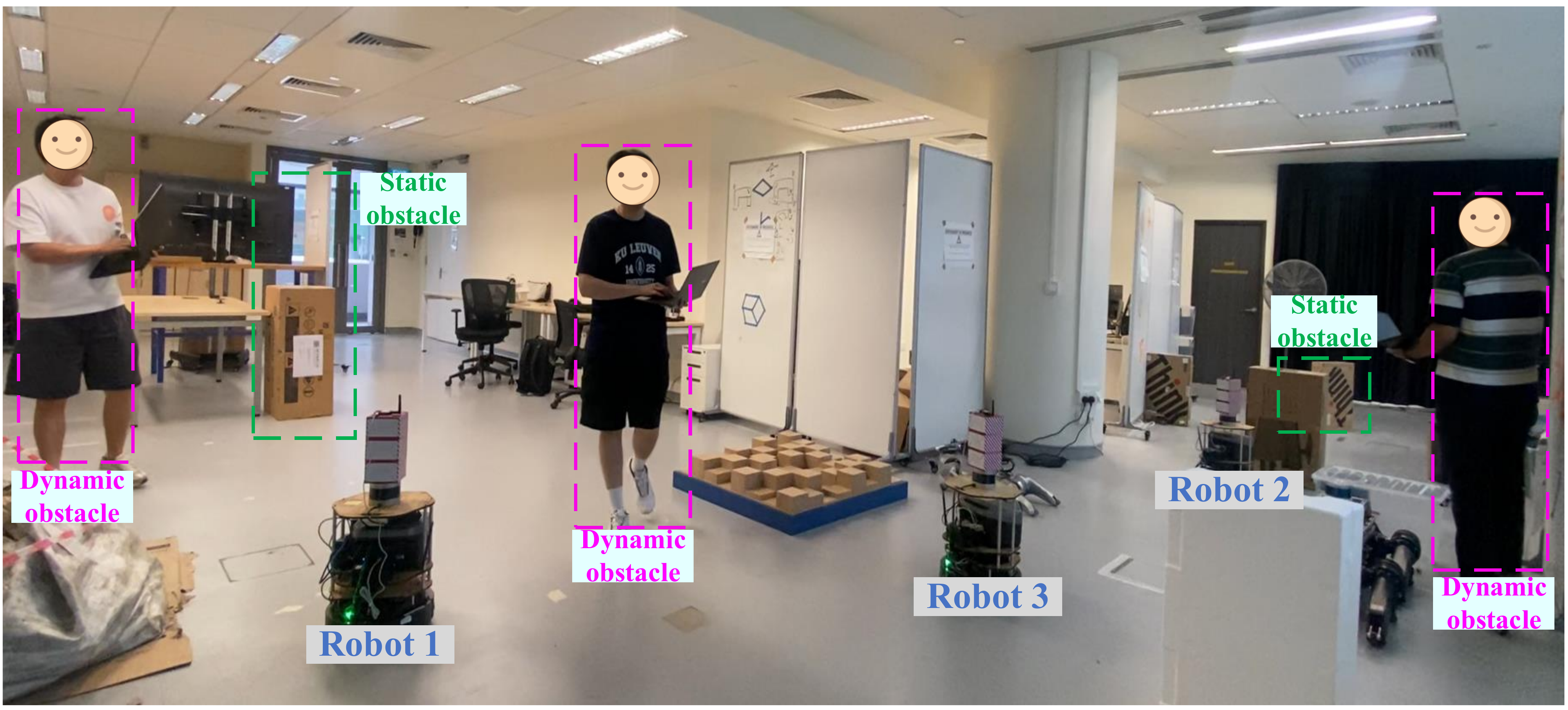}
    \caption{\small Three mobile robots moved randomly in a 14.2m$\times$14.2m unstructured environment containing dynamic and static obstacles.}
    \label{figure-complex-environment}
\end{figure}

\section{Discussion}\label{section-discussion}
In prior experiment, the proposed approach achieved high accuracy and stability in a structured environment.
Here, to more comprehensively evaluate its practicality and reliability, we investigate its performance in both complex and unstructured scenarios and discuss its scalability through simulations.

\begin{figure}[t]
\centering
\setlength{\abovecaptionskip}{0pt}%
\setlength{\belowcaptionskip}{0pt}%

% ----- Column 1 -----
\subfloat[\footnotesize Robot 1 perspective]{%
    \begin{minipage}[t]{0.16\textwidth}
        \centering
        \includegraphics[width=\linewidth]{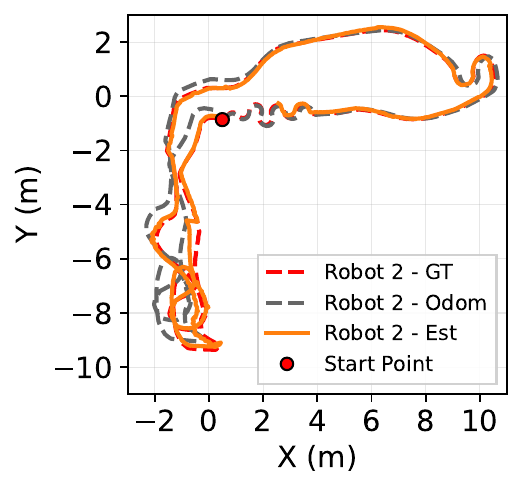}\\[-0.3em]
        \includegraphics[width=\linewidth]{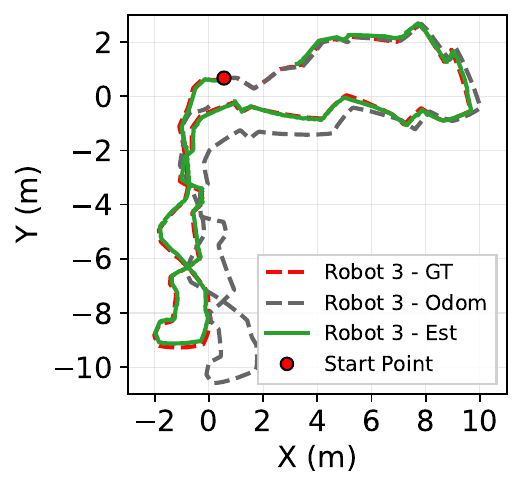}
    \end{minipage}%
    \label{fig:col1}
}
\hspace{-0.4em}%
% ----- Column 2 -----
\subfloat[\footnotesize Robot 2 perspective]{%
    \begin{minipage}[t]{0.16\textwidth}
        \centering
        \includegraphics[width=\linewidth]{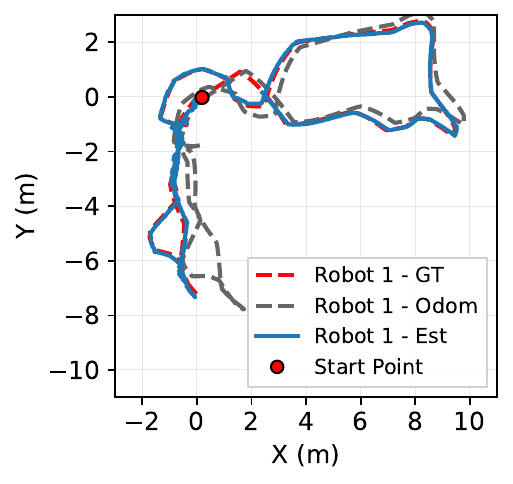}\\[-0.3em]
        \includegraphics[width=\linewidth]{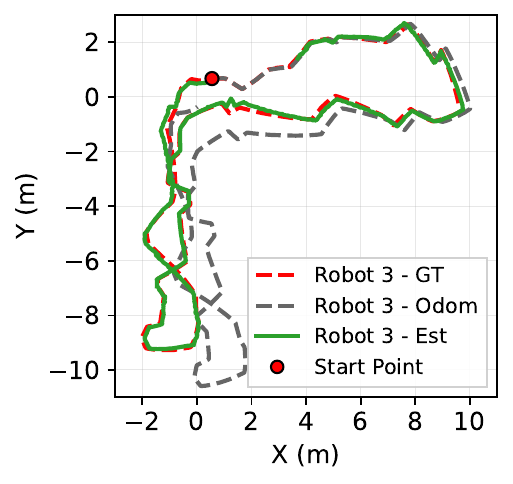}
    \end{minipage}%
    \label{fig:col2}
}
\hspace{-0.4em}%
% ----- Column 3 -----
\subfloat[\footnotesize Robot 3 perspective]{%
    \begin{minipage}[t]{0.16\textwidth}
        \centering
        \includegraphics[width=\linewidth]{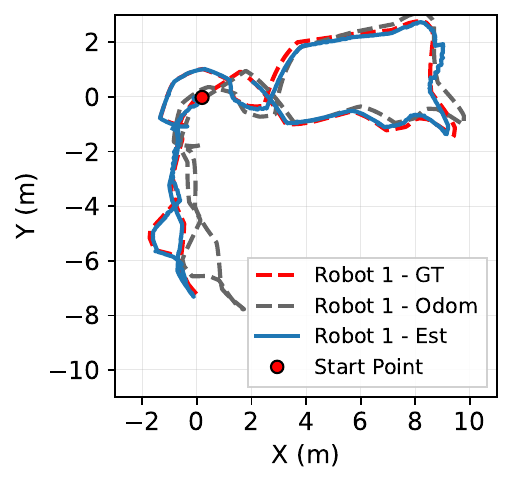}\\[-0.3em]
        \includegraphics[width=\linewidth]{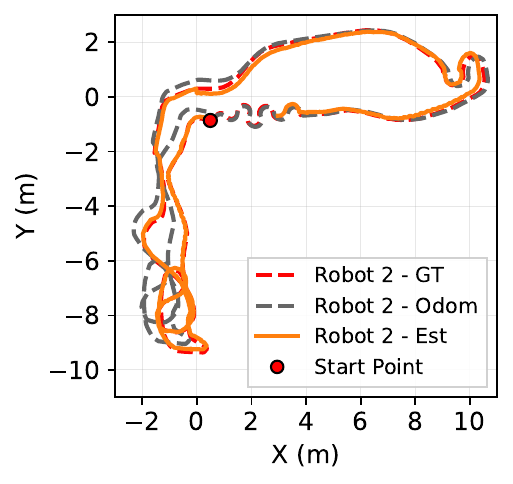}
    \end{minipage}%
    \label{fig:col3}
}

\caption{\small Visualization of pose estimation from different observing robots' viewpoints. Trajectories are obtained by transferring the estimated relative pose to the ground truth of the observing robot.}
\label{figure-complex-env-trajectory}
\end{figure}

\subsection{Evaluation in a Complex Environment}
We conducted the experiment in an unstructured environment with a size of 14.2m$\times$14.2m, using three mobile robots configured similarly to those described in Section \ref{subsection-experimental-setup}.
Fig. \ref{figure-complex-environment} depicts the experimental scenario, including various static obstacles, randomly moving pedestrians, and mobile robots.
The experiment was conducted over a duration of 360 seconds, during which the robots encountered potential inference from both static obstacles and moving pedestrians.
In the last 30 seconds, Robot 1 first returned to its initial position and remained stationary, while Robot 2 and Robot 3 continued moving.
The trajectories of pose estimation are visualized in Fig. \ref{figure-complex-env-trajectory}.
The observing robot is able to continuously localize its teammate robots.
The proposed approach obtained an average translational error of 0.11m and rotational error of 2.12$^{\circ}$.
The results indicate that the performance in the complex environment is consistent with that obtained in the previous experiment conducted in Section \ref{section-experiments}, showing the reliability of the proposed approach.
Due to space limitations, additional performance details are provided in the attached video.

\subsection{Scalability Analysis}
\begin{figure}[t]
\centering
\vspace{-\baselineskip}
\setcounter{subfigure}{0}
\setlength{\abovecaptionskip}{0pt}%    
\setlength{\belowcaptionskip}{-6pt}%
\subfloat[Simulation environment]{
    \includegraphics[width=39mm]{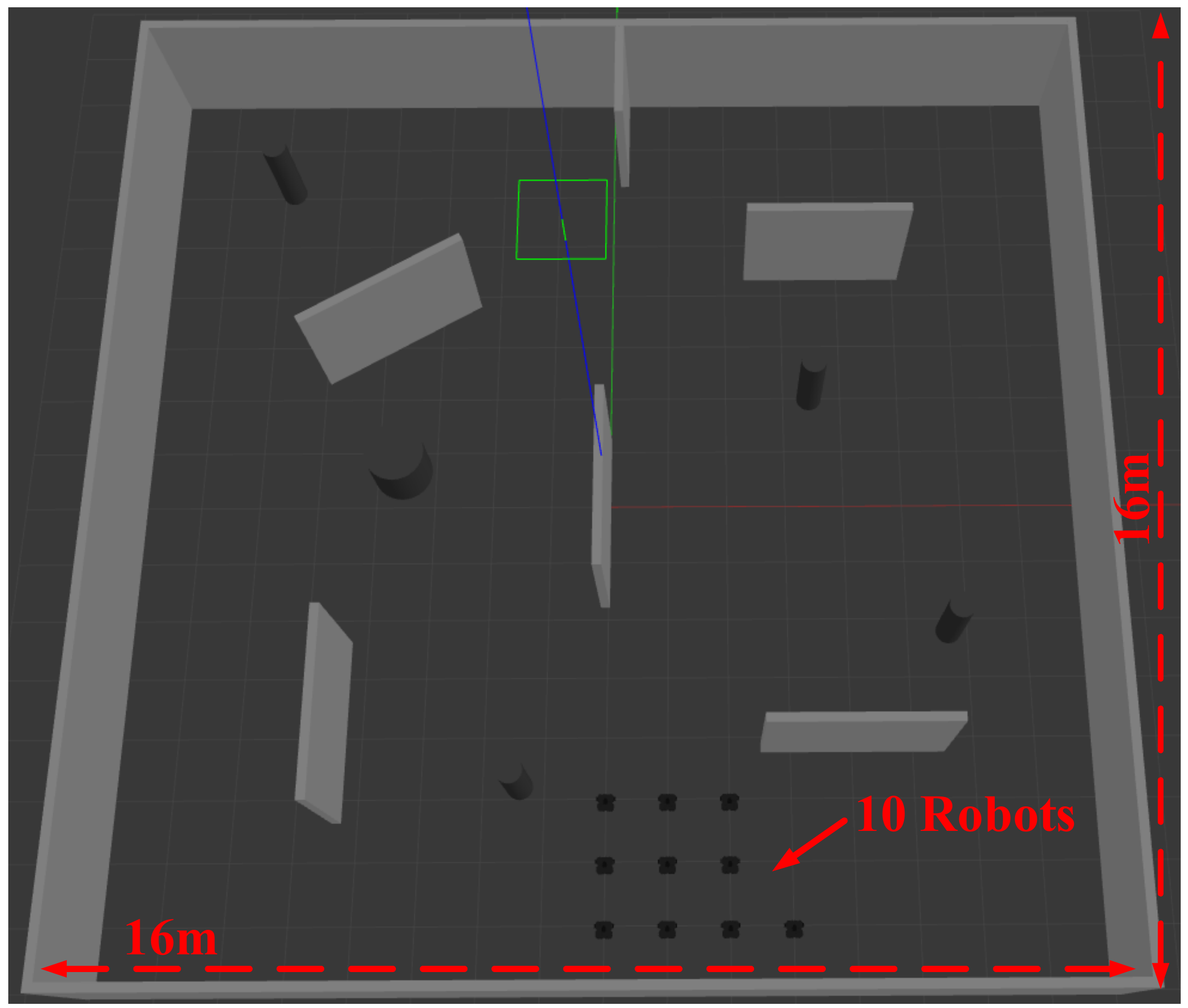}
    \label{scalability-exp-env}
}
\subfloat[Ground truth and odometry trajectory]{
    \includegraphics[width=47mm]{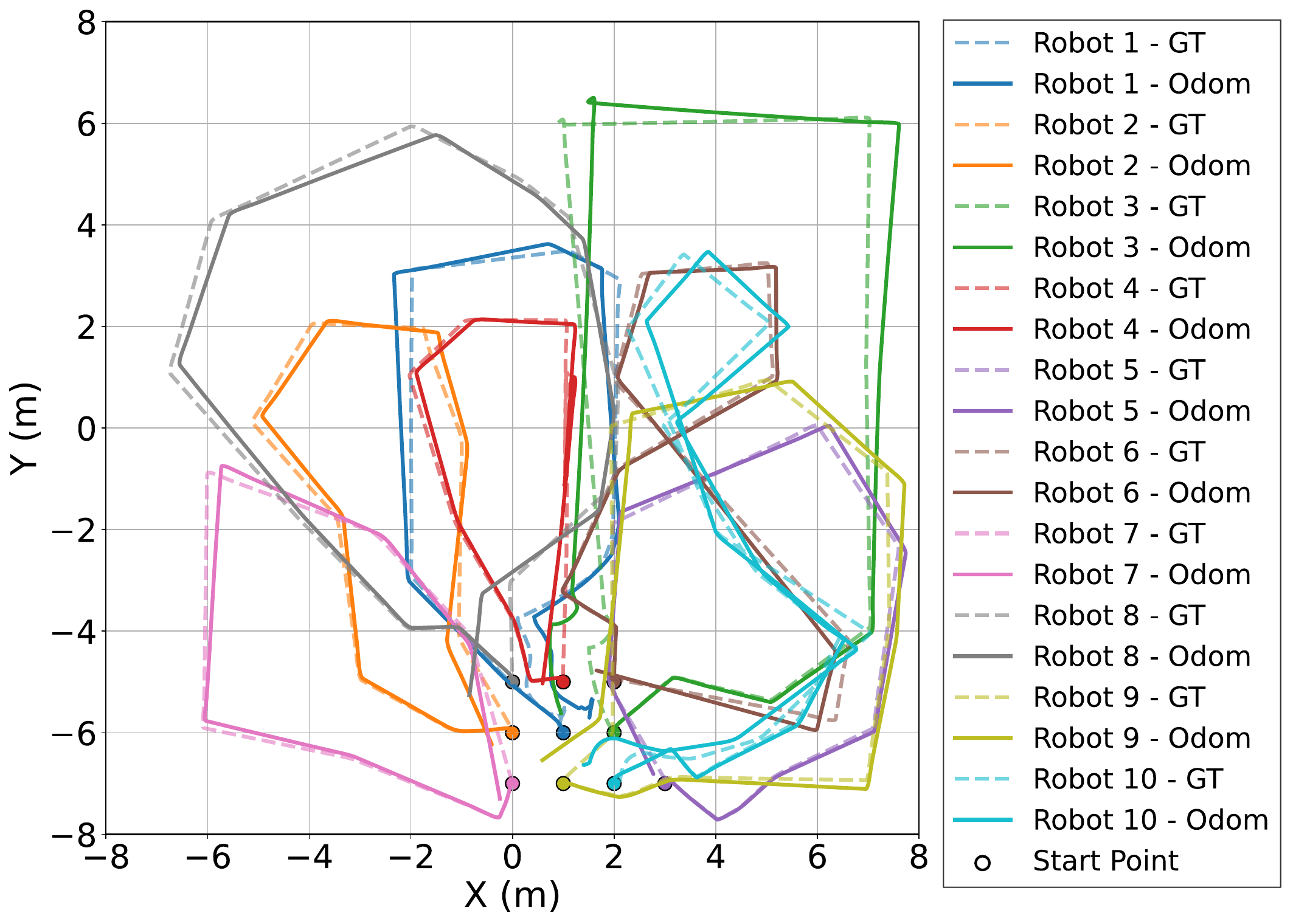}
    \label{scalability-gt-odom}
}
\caption{\small Simulation environment and trajectories of ten robots. (a) Gazebo-based simulation environment with a size of 16m$\times$16m. (b) Ground truth and odometry trajectories.}
\label{figure-scalability-simulation-setting}
\end{figure}

To discuss the scalability of the proposed approach, we conducted simulations with different team sizes varying from 3 to 10 identically configured TurtleBot3 robots.
All trials were performed in the same Gazebo environment to ensure fair comparison, as shown in Fig. \ref{figure-scalability-simulation-setting}.
Table \ref{table-scalability} lists the relative localization accuracy for four experimental trials, with team sizes of 3, 5, 7, and 10 robots, respectively.
Due to space limitations, only results from Robot 1's perspective are presented.
The results indicate that the proposed approach scales effectively to a 10-robot team while maintaining consistent relative localization accuracy with that observed in smaller teams.
However, scaling the team size from 3 to 10 robots increases communication bandwidth usage from 0.48KB/s to 1.6KB/s and prolongs the average time consumption of teammate identification from 6.3ms to 66.7ms.
The communication bandwidth usage increases primarily due to the larger number of robots participating in the data exchange, as calculated from the formulation in Section \ref{subsection-experiment-communication}.
Similarly, the rise in teammate identification time results from the greater number of candidate associations that must be processed during the identification procedure with a larger team size.
Nevertheless, these increases remain acceptable.
Moreover, the performance of other modules is minimally impacted, benefiting from the system’s distributed design.

\begin{table}[t]
\centering
\vspace{-\baselineskip}
\setlength{\abovecaptionskip}{0pt}%    
\setlength{\belowcaptionskip}{0pt}%
\caption{\small Relative pose estimation accuracy (i.e., translational error and rotational error) of our approach from Robot 1's perspective. Trials 1–4 correspond to teams of 3, 5, 7, and 10 robots, respectively. '-' denotes robot absent from the trial.}
\label{table-scalability}
\resizebox{.49\textwidth}{!}
{
\begin{tabular}{|c|ccccc|}
\hline
\multirow{2}{*}{}                                                                                   & \multicolumn{5}{c|}{\begin{tabular}[c]{@{}c@{}}Average   relative translational error (in meters) and \\ rotational error (in degrees) with respect to Robot 1\end{tabular}} \\ \cline{2-6} 
                                                                                                    & \multicolumn{1}{c|}{}                       & \multicolumn{1}{c|}{Trial 1}        & \multicolumn{1}{c|}{Trial 2}        & \multicolumn{1}{c|}{Trial 3}       & Trial 4       \\ \hline
\multirow{2}{*}{Robot 2}                                                                            & \multicolumn{1}{c|}{Trans. error}           & \multicolumn{1}{c|}{0.09±0.05}      & \multicolumn{1}{c|}{0.11±0.09}      & \multicolumn{1}{c|}{0.11±0.09}     & 0.14±0.09     \\ \cline{2-6} 
                                                                                                    & \multicolumn{1}{c|}{Rot. error}             & \multicolumn{1}{c|}{2.86±2.58}      & \multicolumn{1}{c|}{2.34±1.64}      & \multicolumn{1}{c|}{2.48±1.97}     & 2.01±1.35     \\ \hline
\multirow{2}{*}{Robot 3}                                                                            & \multicolumn{1}{c|}{Trans. error}           & \multicolumn{1}{c|}{0.18±0.12}      & \multicolumn{1}{c|}{0.16±0.11}      & \multicolumn{1}{c|}{0.16±0.10}     & 0.21±0.13     \\ \cline{2-6} 
                                                                                                    & \multicolumn{1}{c|}{Rot. error}             & \multicolumn{1}{c|}{3.10±2.01}      & \multicolumn{1}{c|}{2.78±1.88}      & \multicolumn{1}{c|}{2.78±1.60}     & 3.47±2.10     \\ \hline
\multirow{2}{*}{Robot 4}                                                                            & \multicolumn{1}{c|}{Trans. error}           & \multicolumn{1}{c|}{-}              & \multicolumn{1}{c|}{0.08±0.05}      & \multicolumn{1}{c|}{0.08±0.05}     & 0.08±0.04     \\ \cline{2-6} 
                                                                                                    & \multicolumn{1}{c|}{Rot. error}             & \multicolumn{1}{c|}{-}              & \multicolumn{1}{c|}{1.48±1.11}      & \multicolumn{1}{c|}{1.59±1.32}     & 2.77±2.67    \\ \hline
\multirow{2}{*}{Robot 5}                                                                            & \multicolumn{1}{c|}{Trans. error}           & \multicolumn{1}{c|}{-}              & \multicolumn{1}{c|}{0.13±0.06}      & \multicolumn{1}{c|}{0.11±0.05}     & 0.11±0.04     \\ \cline{2-6} 
                                                                                                    & \multicolumn{1}{c|}{Rot. error}             & \multicolumn{1}{c|}{-}              & \multicolumn{1}{c|}{2.35±1.43}      & \multicolumn{1}{c|}{2.42±1.54}     & 2.49±1.29     \\ \hline
\multirow{2}{*}{Robot 6}                                                                            & \multicolumn{1}{c|}{Trans. error}           & \multicolumn{1}{c|}{-}              & \multicolumn{1}{c|}{-}              & \multicolumn{1}{c|}{0.10±0.06}     & 0.13±0.06     \\ \cline{2-6} 
                                                                                                    & \multicolumn{1}{c|}{Rot. error}             & \multicolumn{1}{c|}{-}              & \multicolumn{1}{c|}{-}              & \multicolumn{1}{c|}{1.75±1.07}     & 2.26±1.44     \\ \hline
\multirow{2}{*}{Robot 7}                                                                            & \multicolumn{1}{c|}{Trans. error}           & \multicolumn{1}{c|}{-}              & \multicolumn{1}{c|}{-}              & \multicolumn{1}{c|}{0.11±0.10}     & 0.09±0.06     \\ \cline{2-6} 
                                                                                                    & \multicolumn{1}{c|}{Rot. error}             & \multicolumn{1}{c|}{-}              & \multicolumn{1}{c|}{-}              & \multicolumn{1}{c|}{1.90±1.60}     & 2.39±1.86     \\ \hline
\multirow{2}{*}{Robot 8}                                                                            & \multicolumn{1}{c|}{Trans. error}           & \multicolumn{1}{c|}{-}              & \multicolumn{1}{c|}{-}              & \multicolumn{1}{c|}{-}             & 0.07±0.04     \\ \cline{2-6} 
                                                                                                    & \multicolumn{1}{c|}{Rot. error}             & \multicolumn{1}{c|}{-}              & \multicolumn{1}{c|}{-}              & \multicolumn{1}{c|}{-}             & 1.49±1.07     \\ \hline
\multirow{2}{*}{Robot 9}                                                                            & \multicolumn{1}{c|}{Trans. error}           & \multicolumn{1}{c|}{-}              & \multicolumn{1}{c|}{-}              & \multicolumn{1}{c|}{-}             & 0.12±0.06     \\ \cline{2-6} 
                                                                                                    & \multicolumn{1}{c|}{Rot. error}             & \multicolumn{1}{c|}{-}              & \multicolumn{1}{c|}{-}              & \multicolumn{1}{c|}{-}             & 2.88±1.76     \\ \hline
\multirow{2}{*}{Robot 10}                                                                           & \multicolumn{1}{c|}{Trans. error}           & \multicolumn{1}{c|}{-}              & \multicolumn{1}{c|}{-}              & \multicolumn{1}{c|}{-}             & 0.12±0.06     \\ \cline{2-6} 
                                                                                                    & \multicolumn{1}{c|}{Rot. error}             & \multicolumn{1}{c|}{-}              & \multicolumn{1}{c|}{-}              & \multicolumn{1}{c|}{-}             & 2.41±1.66     \\ \hline
\multirow{2}{*}{\textbf{\begin{tabular}[c]{@{}c@{}}Mean error\\ across robots\end{tabular}}} & \multicolumn{1}{c|}{\textbf{Trans. error}}  & \multicolumn{1}{c|}{\textbf{0.14}}  & \multicolumn{1}{c|}{\textbf{0.12}}  & \multicolumn{1}{c|}{\textbf{0.11}} & \textbf{0.12} \\ \cline{2-6} 
                                                                                                    & \multicolumn{1}{c|}{\textbf{Rot. error}}    & \multicolumn{1}{c|}{\textbf{2.98}}  & \multicolumn{1}{c|}{\textbf{2.24}}  & \multicolumn{1}{c|}{\textbf{2.15}} & \textbf{2.46} \\ \hline
\end{tabular}
}
\end{table}

\vspace{-8pt}
\section{Conclusion and Future Works}\label{section-conclusion-and-future-works}
In this paper, we present a distributed relative localization method for a group of homogeneous robots using only onboard UWB communication network, enabling flexible and robust multi-robot cooperation in communication-limited scenarios, such as post-disaster search and rescue or unknown-environment exploration, even without prior knowledge of initial correspondence among robots.
Particularly, we identify teammate robots from the tracked anonymous LiDAR clusters by a joint matching strategy.
Subsequently, a teammate tracker is employed to continuously localize the teammate robot in a distributed manner.
The robustness and feasibility of the proposed approach were validated through experiments with three homogeneous robots in an indoor environment.
The results demonstrate that our approach achieves an average relative translational error of approximately 0.12m and relative rotational error of approximately 2.20$^{\circ}$, with data exchange of only 0.48KB/s at a communication frequency of 20Hz in a three-robot system, while maintaining robust performance.
%The average packet loss rate of the UWB communication network is only 0.467\% across three robots.
In future work, we will extend our method to 3D scenarios by replacing wheel odometry with an IMU to enable 3D ego-state estimation while enhancing scalability for larger multi-robot systems.
Moreover, we aim to apply our relative localization approach to support the exploration and formation tasks for robot swarm in both indoor and outdoor environments.

\vspace{-1.4cm}

\begin{IEEEbiography}[\vspace{-5pt}{\includegraphics[width=1.0in,height=1.2in,clip,keepaspectratio]{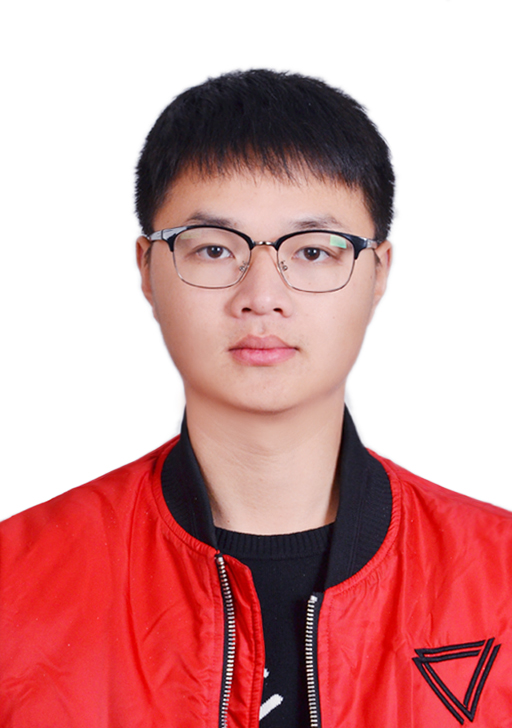}}]{Zhiqiang Cao}
received the B.S. and M.A.Sc. degrees from the Southwest University of Science and Technology, Mianyang, China, in 2019 and 2022, respectively. He is currently pursuing his Ph.D. degree at the Singapore University of Technology and Design, Singapore, under the supervision of Prof. Chau Yuen. His current research interests include multi-robot systems, collaborative localization, distributed systems, and data fusion.
\end{IEEEbiography}

\vspace{-1.7cm}

\begin{IEEEbiography}[{\includegraphics[width=1.0in,height=1.2in,clip,keepaspectratio]{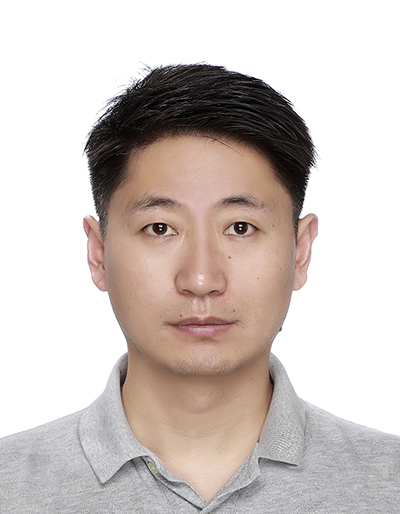}}]{Ran Liu}
(Senior Member, IEEE) received the B.S. degree from the Southwest University of Science and Technology, China, in 2007, and the Ph.D. degree from the University of Tüebingen, Germany, in 2014. From 2015 to 2024, he was a research fellow at the Singapore University of Technology and Design. He is currently a senior research fellow at the Nanyang Technological University. His research interests include robotics and SLAM.
\end{IEEEbiography}

\vspace{-1.6cm}

\begin{IEEEbiography}[\vspace{10pt}{\includegraphics[width=1in,height=1.2in,clip,keepaspectratio]{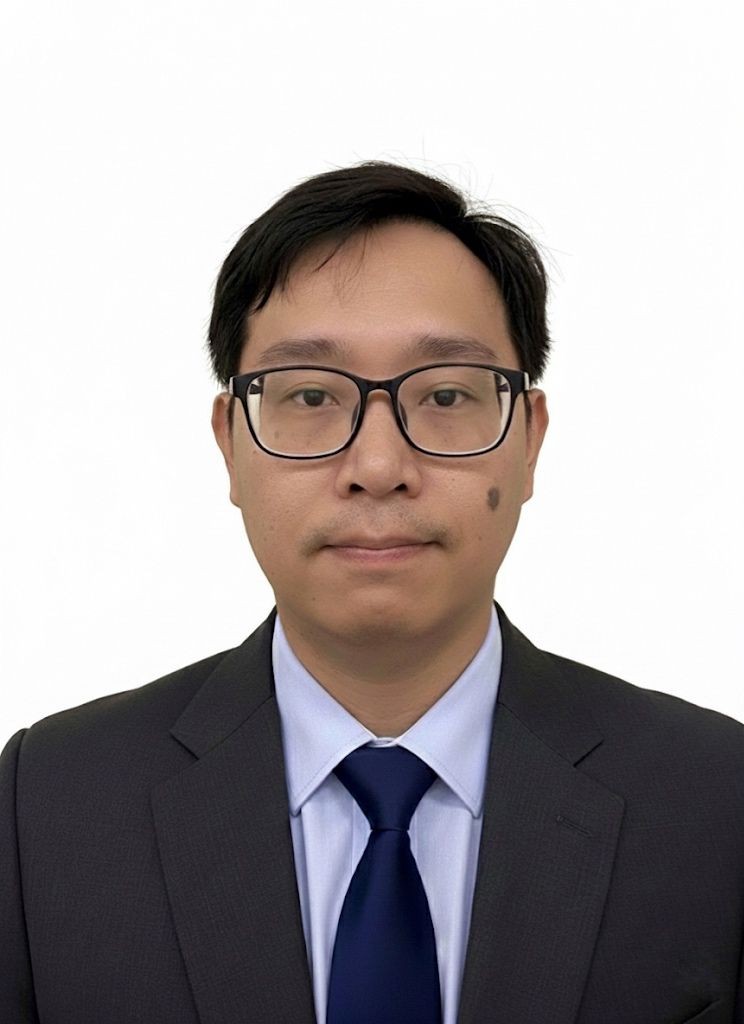}}]{Billy Pik Lik Lau}
(Senior Member, IEEE) received Ph.D. at the Singapore University of Technology and Design (SUTD) in 2021 as well as a degree in computer science and M. Phil degree in computer science from Curtin University in 2010 and 2014, respectively.
His Ph.D. research work includes data fusion and big data analysis in field such as urban science and Internet of Things.
Currently, he is a research fellow at Nanyang Technological University doing data fusion with robotic application.
His research interests are the integration of data fusion and data analysis with multi-agent system for real-world applications.
\end{IEEEbiography}

\vspace{-1.3cm}

\begin{IEEEbiography}[{\includegraphics[width=1in,height=1.2in,clip,keepaspectratio]{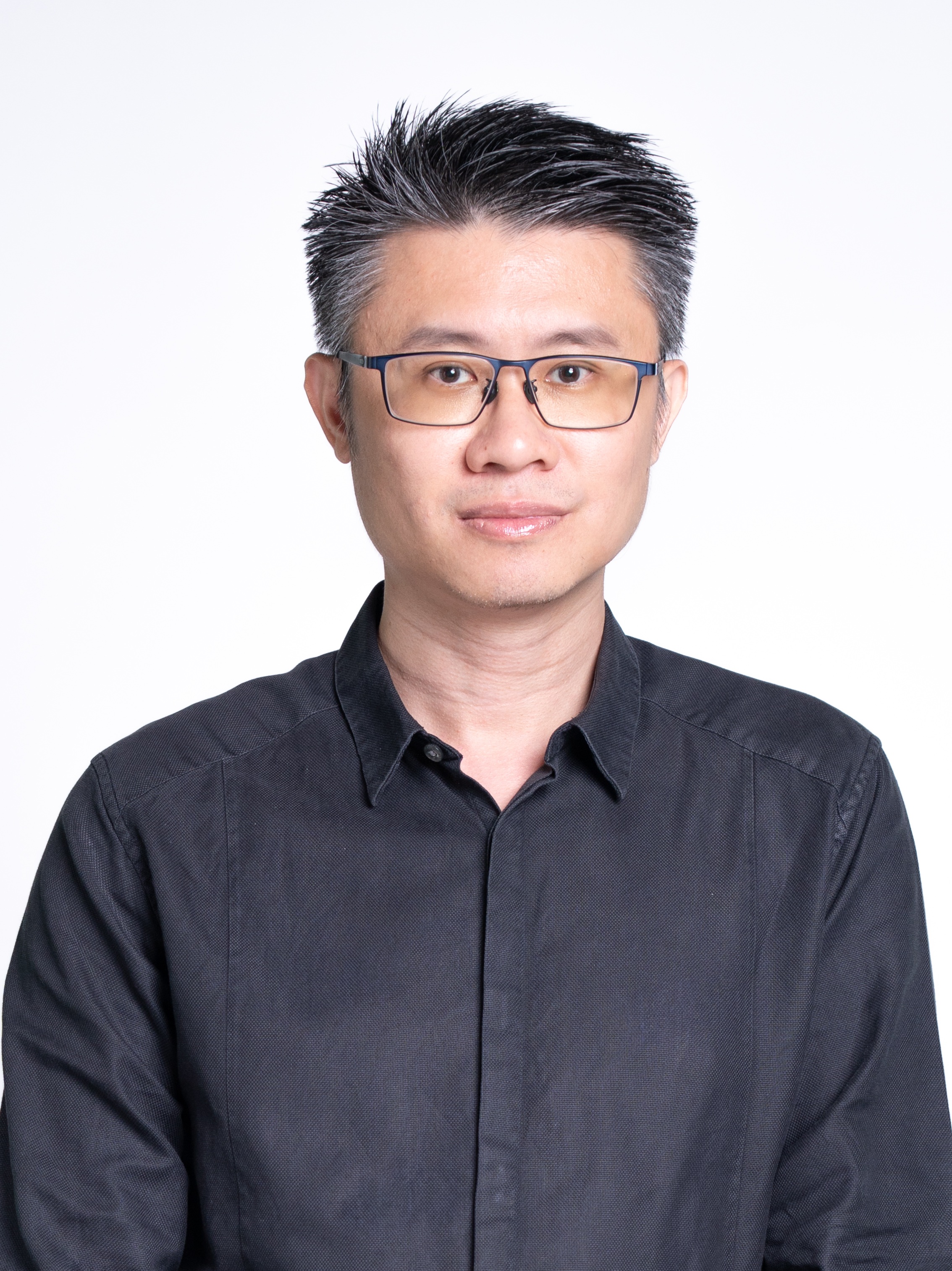}}]{Chau Yuen}
(Fellow, IEEE) received the B.Eng. and Ph.D. degrees from Nanyang Technological University, Singapore, in 2000 and 2004, respectively.
He was a Post-Doctoral Fellow with Lucent Technologies Bell Labs, Murray Hill, in 2005. From 2006 to 2010, he was with the Institute for Infocomm Research, Singapore. From 2010 to 2023, he was with the Engineering Product Development Pillar, Singapore University of Technology and Design. Since 2023, he has been with the School of Electrical and Electronic Engineering, Nanyang Technological University, currently he is Provost’s Chair in Wireless Communications and Assistant Dean in Graduate College.
He is listed as Top 2\% Scientists by Stanford University, and also a Highly Cited Researcher by Clarivate Web of Science from 2022.
Dr. Yuen received IEEE Communications Society Leonard G. Abraham Prize (2024), IEEE Communications Society Best Tutorial Paper Award (2024).
Dr. Yuen currently serves as an Editor-in-Chief for Springer Nature Computer Science, Editor for IEEE Transactions on Vehicular Technology, IEEE Transactions on Neural Networks and Learning Systems, and IEEE Transactions on Network Science and Engineering.
\end{IEEEbiography}

\vspace{-1.3cm}

\begin{IEEEbiography}[{\includegraphics[width=1in,height=1.2in,clip,keepaspectratio]{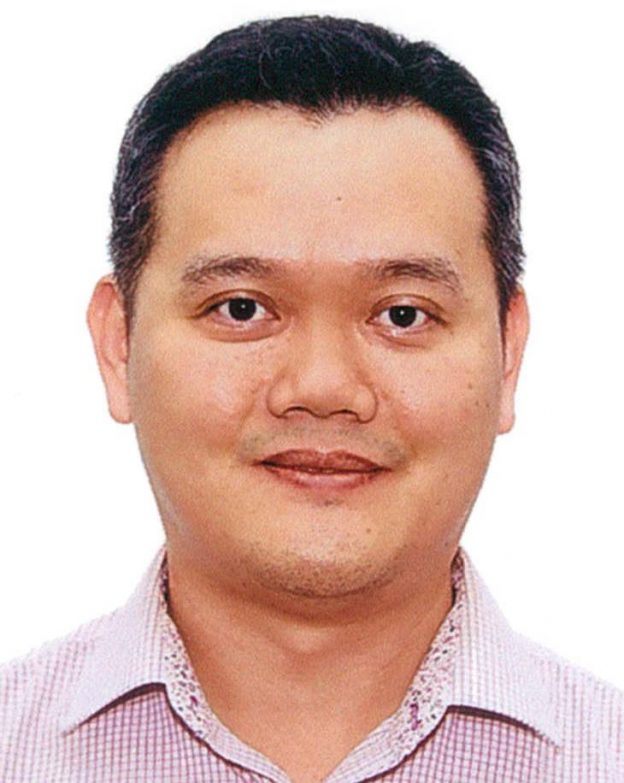}}]{U-Xuan Tan}
(Senior Member, IEEE) received the B.Eng. degree in mechanical and aerospace engineering and the Ph.D. degree in mechatronics from Nanyang Technological University, Singapore, in 2005 and 2010, respectively. From 2009 to 2011, he was Post-Doctoral Researcher at the University of Maryland, College Park, MD, USA. From 2012 to 2014, he was a Lecturer with Singapore University of Technology and Design (SUTD), Singapore, where he took up a research intensive role as an Assistant Professor in 2014, and has been promoted to an Associate Professor since 2021. He is currently the Chair of the SUTD Institutional Review Board and is also a Senior Visiting Academician with the Changi General Hospital. His research interests include Robotics \& Automation, onsite sensing and control, SLAM in adverse conditions, sensing algorithms for human–robot interaction, and interdisciplinary teaching.
\end{IEEEbiography}

\end{document}